\documentclass{article} 
\usepackage{iclr2027_conference,times}

\usepackage{amsmath,amsfonts,bm}

\newcommand{\Ldp}{\mathcal{L}_{\textsc{dp}}}

\def\eqref#1{equation~\ref{#1}}

\def\1{\bm{1}}

\def\eps{{\epsilon}}

\DeclareMathAlphabet{\mathsfit}{\encodingdefault}{\sfdefault}{m}{sl}
\SetMathAlphabet{\mathsfit}{bold}{\encodingdefault}{\sfdefault}{bx}{n}

\newcommand{\E}{\mathbb{E}}

\newcommand{\R}{\mathbb{R}}

\usepackage{hyperref}
\usepackage{titletoc}
\usepackage{url}
\usepackage[most]{tcolorbox}
\usepackage{wrapfig}
\usepackage{xcolor}
\usepackage{capt-of}
\usepackage{booktabs}
\usepackage{wrapfig}
\usepackage{amsmath}
\usepackage{caption}
\usepackage{graphicx}
\usepackage{lipsum}
\numberwithin{equation}{section}
\newif\ifshowcomments
\showcommentstrue        

\definecolor{selectedorange}{HTML}{E08262}
\definecolor{finalblue}{HTML}{5E93CE}

\usepackage{color-edits}
\addauthor{ms}{magenta}

\definecolor{takeawayviolet}{HTML}{70519B}
\definecolor{takeawayvioletfill}{HTML}{F3EFF8}

\newcommand{\methodfont}[1]{{\texttt{\textbf{#1}}}}

\definecolor{baselinegraytext}{HTML}{747474}
\definecolor{jointpurpletext}{HTML}{594388}
\definecolor{postbcbluetext}{HTML}{5EAED6}
\definecolor{tmrlambertext}{HTML}{C9A227}

\newcommand{\BaselineDP}{\texorpdfstring{{\color{baselinegraytext}\methodfont{Baseline DP}}}{Baseline DP}}
\newcommand{\JointOA}{\texorpdfstring{{\color{jointpurpletext}\methodfont{JointAug}}}{JointAug}}
\newcommand{\PostBC}{\texorpdfstring{{\color{postbcbluetext}\methodfont{PostBC}}}{PostBC}}
\newcommand{\TMRL}{\texorpdfstring{{\color{tmrlambertext}\methodfont{TMRL}}}{TMRL}}

\definecolor{denseblue}{HTML}{2A7FB8}
\definecolor{mediumgray}{HTML}{7F8C8D}
\definecolor{sparsered}{HTML}{C0392B}

\newtcolorbox[auto counter]{takeawaybox}{
    enhanced,
    breakable,
    colback=takeawayvioletfill,
    colframe=takeawayviolet,
    boxrule=0.8pt,
    arc=1.5mm,
    left=2.5mm,
    right=2.5mm,
    top=1.5mm,
    bottom=1.5mm,
    before skip=8pt,
    after skip=8pt,
    before upper={\textbf{Takeaway~\thetcbcounter:}\ }
}

\newcommand{\appsubsec}[1]{%
    \refstepcounter{subsection}%
    \addcontentsline{toc}{subsection}{%
        \protect\numberline{\thesubsection}#1}%
    \par\medskip
    \noindent
    {\hyphenpenalty=10000\exhyphenpenalty=10000
     \textbf{#1.}}%
    \nobreak\hspace{0.4em}\ignorespaces
}

\definecolor{softlinkblue}{HTML}{6BA7D6}
\definecolor{citationblue}{HTML}{1F4E79}

\hypersetup{
    colorlinks=true,
    linkcolor=citationblue,
    citecolor=citationblue,
    urlcolor=softlinkblue
}

\title{Beyond Validation Loss: \\ Coverage and Reliability Dynamics in Robot Policy Pretraining}

\author{%
  {\fontsize{10pt}{12pt}\selectfont\bfseries
   Zhanyi Sun\textsuperscript{1}\quad
   Max Simchowitz\textsuperscript{2,*}\quad
   Shuran Song\textsuperscript{1,*}}\\[0.5em]
  {\normalfont\fontsize{9pt}{10.8pt}\selectfont
   \textsuperscript{1}Stanford University\qquad
   \textsuperscript{2}Carnegie Mellon University}\\[0.05em]
  {\normalfont\fontsize{9pt}{10.8pt}\selectfont
   \textsuperscript{*}Equal advising.\quad
   Correspondence to: \texttt{zhanyis@stanford.edu}}
}

\iclrfinalcopy 
\begin{document}

\maketitle
\fancyhead[L]{Preprint}

\begin{abstract}
In modern robot learning,  generative robot policies are first pretrained through supervised learning from demonstrations, then refined through reinforcement learning, supervised finetuning, or other forms of posttraining.
Yet conventional pretraining metrics, such as validation loss and average task success, provide an incomplete account of which policies support both effective direct deployment and subsequent posttraining.
In this work, we study closed-loop pass@$k$, which evaluates the probability that $k$ rollouts from the same initial condition attain at least one success. 
Thus, pass@$1$ measures mean task success; large-$k$ pass@$k$ measures whether success remains reachable under repeated sampling. 
Across a wide range of posttraining recipes, we show that large-$k$ pass@$k$ reveals posttraining differences not captured by pass@1. In particular, checkpoints with similar pass@$1$ but different large-$k$ pass@$k$ have different posttraining performance.
Moreover, we show pass@$k$ peaks at different checkpoints during supervised pretraining, revealing a coverage--reliability tradeoff over the course of training. Based on these observations, we propose two practical interventions: (1) a purely \emph{offline} checkpoint-selection criterion that balances coverage and reliability without closed-loop evaluation, and (2) a generative data augmentation procedure that expands supervision beyond the finite demonstration set, preserving large-$k$ pass@$k$ while maintaining or improving pass@$1$ and raising the observed coverage--reliability Pareto frontier. Together, our results motivate evaluation, checkpoint selection, and training procedures that explicitly account for posttraining. Blog post, visualizations, and code are available at
\href{https://zhanyisun.github.io/robot-policy-learning-dynamics/}{zhanyisun.github.io/robot-policy-learning-dynamics/}.

\end{abstract}

\section{Introduction}
\label{sec:intro}

Robot policies are increasingly trained in stages: supervised behavior cloning (BC) or large-scale pretraining \citep{chi2025diffusion, kim2024openvla, black2024pi_0}, followed by reinforcement learning (RL), supervised finetuning (SFT), or other forms of adaptation.
Yet supervised pretraining and posttraining are usually studied as separate problems.
The checkpoint handed to posttraining is typically chosen by validation loss, average task success, or simply taking the final checkpoint --- criteria that evaluate the supervised pretraining stage in isolation.
This leaves a central question unresolved: \emph{how do the learning dynamics of supervised pretraining shape both direct deployment and downstream finetunability, and what criterion can identify checkpoints that jointly support both?}

Language-model research provides strong motivation to look beyond metrics and progress within a single stage of training in isolation. In particular, models with similar pretraining validation loss can perform very differently after posttraining, while prolonged pretraining or stronger SFT performance does not always carry over to subsequent posttraining \citep{liu2023same,springer2025overtrained,kang2026quagmires,zhang2026good}. Recent work further highlights coverage of high-quality outputs as an important property for downstream posttraining \citep{foster2025good,chen2026coverage}.

Robot control, however, presents a different set of challenges than language modeling: policies are densely conditioned on observations, actions are continuous and have far less distributional diversity~\citep{pan2026much}, and success requires stable and precise closed-loop execution~\citep{simchowitz2025pitfalls,zhang2025action}. These demands may appear to favor increasingly reliable policies over broader behavioral coverage.

\textbf{Contributions.} Despite these differences, we find that supervised training of generative robot policies is governed by a similar tradeoff between broad behavioral coverage and reliable task execution. We study this tradeoff through \emph{closed-loop pass@$k$ from a fixed initial condition} (henceforth, fixed-start pass@$k$). For each initial configuration, we rollout the stochastic policy $k$ times and ask whether at least one rollout succeeds, then average over the distribution of initial configurations (Sec.~\ref{sec:problem_setup}). Equivalently, pass@$k$ is the expected probability, over the distribution of initial configurations, that at least one of $k$ independently sampled rollouts from the same configuration succeeds. At $k=1$, this recovers the standard mean task-success rate; we refer to this single-rollout success as \emph{reliability}. 
As $k$ increases, pass@$k$ measures the fraction of initial conditions for which the policy can produce at least one successful rollout within $k$ attempts; we refer to this as \emph{coverage}. The full pass@$k$ profile therefore lets us study how reliability and coverage evolve during pretraining, and how these differences affect posttraining. 

We establish two key findings (Sec.~\ref{sec:passk_dynamics}). \textbf{First}, large-$k$ pass@$k$ tracks downstream posttraining outcomes that pass@$1$ alone doesn't: checkpoints with similar pass@$1$ but different large-$k$ pass@$k$ can yield substantially different performance after subsequent SFT or RL finetuning. \textbf{Second}, reliability and coverage diverge systematically during supervised pretraining. Large-$k$ pass@$k$ peaks earlier and can decline while pass@$1$ continues to improve, revealing that further optimization can improve reliability at the expense of coverage. This tradeoff also matters for direct deployment: declining large-$k$ pass@$k$ reflects the loss of low-probability successful behaviors on harder initial conditions, even as average task success improves.

Motivated by these findings, we propose a novel \textbf{offline criterion} for selecting policy checkpoints that best balance coverage and reliability (Sec.~\ref{sec:offline_diagnosis}). Estimating pass@$k$ requires many closed-loop rollouts from matched initial configurations, which is prohibitively expensive on real robots. Instead, our criterion is computed entirely offline from the quantile profile of held-out action error. This profile captures favorable-tail coverage and typical-case reliability, allowing us to select the checkpoint that best balances the two. Crucially, it requires no closed-loop policy evaluation, no deployment-specific choice of $k$, and no tuned tradeoff weight, making it easy to integrate into modern robot policy training pipelines.

Lastly, we introduce a \textbf{generative data-augmentation procedure} that raises the Pareto frontier between mean task success and large-$k$ pass@$k$ performance (Sec.~\ref{sec:data_aug}). We train a joint observation--action generative model to expand supervision beyond the finite demonstration dataset. The resulting augmentation preserves favorable-tail coverage as pretraining proceeds while maintaining, and in some cases improving, reliability as well. By preserving coverage without sacrificing mean task success, it reduces the need for checkpoint selection and provides stronger initializations for subsequent posttraining.

\section{Problem Setup}
\label{sec:problem_setup}
\textbf{Generative robot policies.}
We study conditional generative policies $\pi_\theta(a\mid o)$ over action chunks, focusing on diffusion-based policies.
Under the noise-prediction parameterization, an expert action chunk $a_0$ is corrupted at noise level $t$, and the denoiser is trained as
\begin{equation}
\label{eq:diffusion_objective}
a_t=\alpha_t a_0+\sigma_t\epsilon,\quad \epsilon\sim\mathcal{N}(0,I),
\qquad
\Ldp(\theta)
=
\mathbb{E}_{(o,a_0),t,\epsilon}
\!\left[
\left\|
\epsilon-\epsilon_\theta(a_t,o,t)
\right\|_2^2
\right].
\end{equation}
At inference, $a_T\sim\mathcal{N}(0,I)$ is iteratively denoised using $\epsilon_\theta$ to produce an action chunk.

\textbf{Pretraining, posttraining, and checkpoint selection.}
We consider robot policies trained through an initial supervised pretraining phase followed by a subsequent posttraining phase.
Along pretraining, optimization produces a sequence of checkpoints ${\pi_\tau}$, where $\tau$ indexes training progress, measured in epochs or optimization steps depending on the training setting.\footnote{Throughout the paper, $\tau$ indexes training progress and should not be confused with the common use of $\tau$ for robot trajectories.}
The selected checkpoint may then be deployed directly or used to initialize posttraining.
In standard practice, checkpoints are commonly selected using validation loss, held-out action MSE, task success rate from closed-loop evaluation, or simply by taking the final checkpoint. Our central question is how the choice of pretraining checkpoint affects performance under direct deployment and after subsequent posttraining, and how to identify checkpoints that support both. 


\textbf{Fixed-start pass@$k$.}
In large language model evaluation on verifiable tasks such as code generation and mathematical reasoning, pass@$k$ is the probability that at least one of $k$ independently sampled solutions is correct~\citep{chen2021evaluating}. For robot policies, we define the analogous closed-loop measure.
For an initial condition $c\sim p_{\mathrm{eval}}$, let $q_\tau(c)$ denote the one-rollout success probability of policy checkpoint $\pi_\tau$.
Then, we define the pass@$k$ for checkpoint $\tau$ as
\begin{equation}
\label{eq:passk}
P_k(\tau)
=
\mathbb{E}_{c\sim p_{\mathrm{eval}}}
\left[
1-\left(1-q_\tau(c)\right)^k
\right].
\end{equation}
That is, for each held-out initial condition, we reset the simulator and execute $k$ independent rollouts from checkpoint $\pi_\tau$; pass@$k$ is the fraction of conditions for which at least one rollout succeeds.
Pass@$1$ recovers the standard mean task-success rate.%
\footnote{We use pass@$k$ as a diagnostic in resettable simulation to study the learning dynamics of pretraining and their relation to finetunability. It is not intended to imply that deployment should require resetting to the same initial condition and executing multiple independent attempts.}

\section{Later Checkpoints Trade Large-\texorpdfstring{$k$}{k} Pass@\texorpdfstring{$k$}{k} for Mean Task Success; Pass@\texorpdfstring{$1$}{1} Can Miss Posttraining Differences}
\label{sec:passk_dynamics}

\begin{figure}[t]
    \centering
    \includegraphics[width=\linewidth]{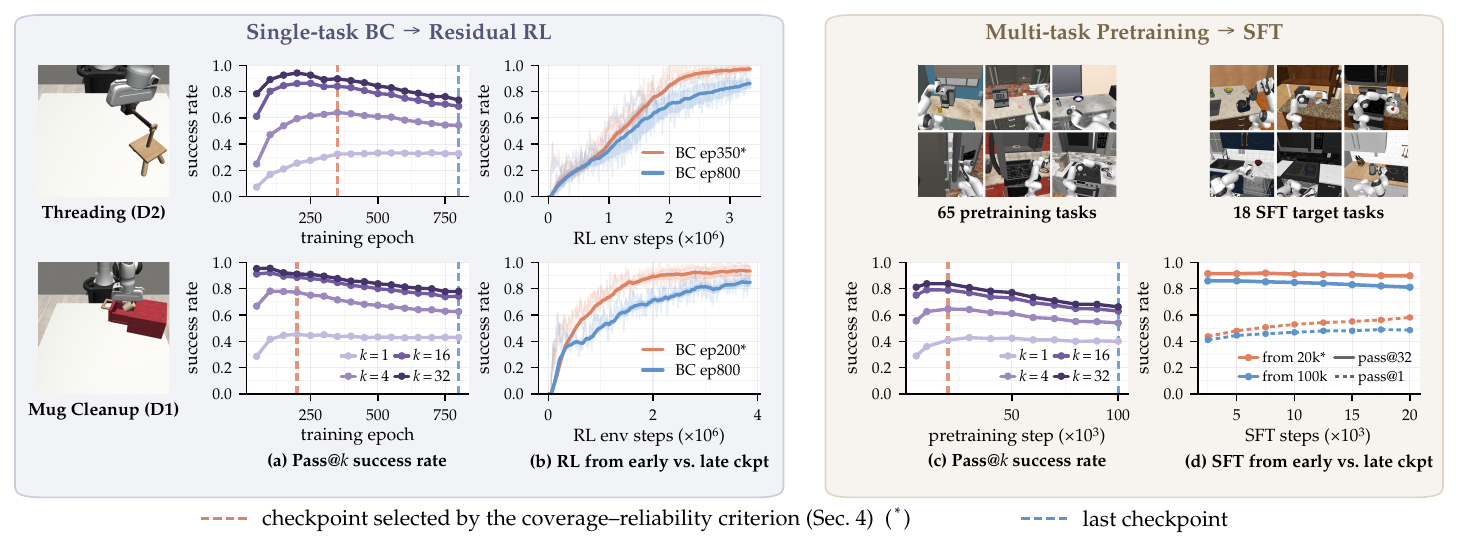}
    \vspace{-0.75cm}
    \caption{\textbf{Later checkpoints trade large-$k$ pass@$k$ for mean task success; checkpoints with similar pass@$1$ can lead to different posttraining performance.}
    \textbf{Left:} single-task BC followed by residual RL on two representative MimicGen tasks, with \textbf{(a)} showing pass@$k$ during BC and \textbf{(b)} residual RL from different BC checkpoints.
    \textbf{Right:} multitask pretraining on RoboCasa tasks followed by SFT on target tasks, with \textbf{(c)} showing pass@$k$ during pretraining and \textbf{(d)} downstream SFT.
    Across \textbf{(a,c)}, larger-$k$ pass@$k$ peaks and declines earlier than pass@$1$, so later checkpoints increasingly favor mean task success at the expense of large-$k$ pass@$k$ performance.
    \textcolor{selectedorange}{Orange} marks the earlier checkpoint selected by our
    \textit{offline coverage--reliability criterion}
    (Sec.~\ref{sec:offline_diagnosis}), while
    \textcolor{finalblue}{blue} marks the final supervised-training checkpoint;
    matching dashed lines in \textbf{(a,c)} indicate the initializations used in
    \textbf{(b,d)}.
    Despite similar pass@$1$, finetuning from the \textcolor{selectedorange}{orange} checkpoint achieves stronger performance than finetuning from the \textcolor{finalblue}{blue} checkpoint across both settings.}
    \label{fig:passk_finetune}
    \vspace{-0.5cm}
\end{figure}

We begin by examining how pass@$k$ curves across $k$ evolve over supervised pretraining, and whether posttraining from checkpoints with similar pass@$1$ but different large-$k$ pass@$k$ leads to different performance.

\textbf{Pretraining evaluation.}
We study supervised-pretraining dynamics in two settings:
(1) single-task BC on MimicGen, and
(2) multitask VLA pretraining on RoboCasa.
For single-task BC, we train image-conditioned Diffusion Policies~\citep{chi2025diffusion}
from scratch on 200 demonstrations for each of five MimicGen tasks~\citep{mandlekar2023mimicgen}.
For multitask VLA pretraining, we use the OpenPI implementation of $\pi_{0.5}$~\citep{intelligence2025pi_}
on 65 atomic RoboCasa365 tasks~\citep{nasiriany2026robocasa365}, with 100 demonstrations per task.
Full evaluation details are provided in App.~\ref{sec:implementation_details}.

For readability, Figure~\ref{fig:passk_finetune}(a,c) shows only two of the five selected MimicGen tasks together with the aggregate RoboCasa results. Main-text MimicGen experiments use 200 demonstrations per task. Later experiments (Sec.~\ref{sec:data_aug}) show all five selected MimicGen tasks and the full RoboCasa evaluation.

\textbf{During supervised pretraining, pass@$k$ peaks earlier as $k$ increases; mean task success favors later checkpoints.}
Figure~\ref{fig:passk_finetune}(a,c) shows a consistent pattern across single-task BC and multitask pretraining: the pass@$k$ optimum shifts \emph{earlier} as $k$ increases. Thus, later checkpoints can achieve higher mean task success even after large-$k$ pass@$k$ has begun to decline. This divergence between mean task success and large-$k$ pass@$k$ reflects \textit{a redistribution of success across conditions} during supervised pretraining.
Recall that $q_\tau(c)$ is the one-rollout success probability from condition $c$. Pass@$1$ depends only on its mean; for $k>1$, we can write $P_k(\tau)=\mathbb{E}_c[f_k(q_\tau(c))]$, where
$f_k(q):=1-(1-q)^k$ is concave in its argument $q$. Shifting success probability from harder conditions toward already-reliable ones can therefore increase average success while decreasing larger-$k$ pass@$k$.
As supervised pretraining proceeds, condition-wise success can become increasingly \emph{polarized}: already-solvable conditions become more reliable while harder conditions lose success probability.
At the behavioral level, we interpret this redistribution as a tradeoff between
\emph{reliability} and \emph{coverage}.
See App.~\ref{app:condition_redistribution} for corresponding condition-level decomposition.

\begin{takeawaybox}
Over supervised pretraining of robot policies, success can become increasingly \textit{polarized} across conditions: mean task success can continue to improve or stabilize as large-$k$ pass@$k$ declines.
\end{takeawaybox}

\textbf{Posttraining evaluation.}
We show that checkpoints with different pass@$k$ profiles can attain substantially different posttraining performance.
For each experiment, we compare the earlier checkpoint selected by our offline coverage--reliability criterion (Sec.~\ref{sec:offline_diagnosis}), called the \textbf{selected} checkpoint, with the \textbf{final} checkpoint from the same supervised pretraining run. The two have similar pass@$1$, but the selected checkpoint retains substantially higher large-$k$ pass@$k$.

We evaluate four posttraining settings.
(1) \emph{Filtered BC}: we collect the same number of rollouts from both checkpoint, retain successful episodes, append them to the original training set, and train a new BC policy.
(2) \emph{steering with a learned critic}: we train one Q-function per checkpoint on the original training data augmented with unfiltered rollouts from that checkpoint.
At held-out state $s$, both critics score the same eight candidate action chunks $\{a_j\}$.
We estimate each candidate's value $p_j$ by branching 16 simulator rollouts from $(s,a_j)$ and measuring the fraction that succeed, then evaluate whether $Q(s,a_j)$ correctly ranks candidates by $p_j$.
(3) \emph{actor-critic residual RL} with DICE-RL~\citep{sun2026prior}, an online RL method for finetuning diffusion
policies.
(4) lastly, on RoboCasa365 only, we perform \emph{supervised finetuning (SFT)} from different pretrained checkpoints using identical target-task finetuning data and optimization.
Full experimental details are in App.~\ref{sec:implementation_details}.

\begin{wraptable}{r}{0.47\columnwidth}
    \centering
    \vspace{-0.6em}
    \footnotesize

    \begin{tabular*}{\linewidth}{@{\extracolsep{\fill}}lcc@{}}
        \toprule
        \textbf{Task} & \textbf{Selected} & \textbf{Final} \\
        \midrule
        Threading (D2)   & \textbf{0.371 / 0.886} & 0.321 / 0.812 \\
        Mug Cleanup (D1) & \textbf{0.558 / 0.914} & 0.502 / 0.866 \\
        \bottomrule
    \end{tabular*}
    \caption{\textbf{Filtered BC.}
    Pass@$1$/pass@$32$ after retraining on successful rollouts generated by the
    selected or final checkpoint.}
    \label{tab:filtered_bc}

    \vspace{0.5em}

    \begin{tabular*}{\linewidth}{@{\extracolsep{\fill}}lcc@{}}
        \toprule
        \textbf{Task} & \textbf{Selected} & \textbf{Final} \\
        \midrule
        Threading (D2)   & \textbf{0.534} & 0.521 \\
        Mug Cleanup (D1) & \textbf{0.589} & 0.547 \\
        \bottomrule
    \end{tabular*}
    \caption{\textbf{Critic action ranking accuracy.}
    Within-state pairwise accuracy against simulator-estimated action values.}
    \label{tab:critic_ranking}

    \vspace{-0.5em}
\end{wraptable}

Together, these experiments provide complementary tests of how pretrained-policy coverage affects posttraining. In \emph{filtered BC}, new supervision comes from successful rollouts~\citep{mirchandani2024so}; under a fixed rollout budget, lower coverage can yield a similar number of successes from fewer conditions, reducing data diversity.
In \emph{critic steering}, broader coverage exposes the critic to a wider range of observation--action outcomes, which can improve generalization when ranking actions at unseen observation; such ranking is central to value-guided test-time steering~\citep{nakamoto2024steering} and actor--critic RL. 
\emph{Actor-critic residual RL} tests whether these differences persist when data collection, critic learning, and policy optimization occur together online. Lastly, \emph{SFT} verifies the findings persist when target-task data are fixed rather than policy-generated rollouts.

\textbf{Checkpoints with large-$k$ pass@$k$ enjoy stronger posttraining performance despite similar pass@$1$ performance.}
Tab.~\ref{tab:filtered_bc} and~\ref{tab:critic_ranking} isolate this effect under \emph{online} data collection.
Across both tasks, retraining on successful rollouts from the selected (higher large-$k$ pass@k) checkpoint produces a stronger filtered BC policy. Indeed, for matched rollout budgets, a greater chance at success on wider variety of initial conditions is more useful than repeated successes on already-solved conditions. Similarly, critic trained on the selected-checkpoint rollouts also rank counterfactual actions more accurately at held-out states, indicating better value-function generalization.

Fig.~\ref{fig:passk_finetune}(b,d) shows that checkpoint selection also affects
actor-critic RL finetuning. Under the same interaction budget, residual RL from the selected checkpoint, which has higher-large-$k$, reaches higher task success than from the final checkpoint. In App.~\ref{app:ogpo}, we also compare different pretrained checkpoints under OGPO, a full-policy RL finetuning method~\citep{patil2026ogpo}. Unlike residual RL, OGPO directly finetunes the pretrained policy rather than learning a residual or steering head on top of a frozen policy. Nevertheless, the selected earlier checkpoint still outperforms the final checkpoint. Lastly, on RoboCasa, SFT from the selected checkpoint is also stronger under identical finetuning data and optimization, showing that the checkpoint effect is not limited to methods trained on policy-generated rollouts.

We note that large-$k$ pass@$k$ does not globally rank checkpoints for finetuning: for much earlier checkpoints with lower pass@$1$, which checkpoint attains higher RL success can depend on finetuning budget and be task-specific. We report results on residual RL finetuning from more pretrained checkpoints in App.~\ref{app:residual_rl_coverage_reliability}. We show that, on some tasks, a much earlier checkpoint with lower pass@$1$ but stronger large-$k$ pass@$k$ can eventually outperform the selected checkpoint given a sufficiently large RL finetuning budget. Among checkpoints with comparable pass@$1$, however, large-$k$ pass@$k$ is a better predictor of the RL posttraining performance. 



\begin{takeawaybox}
Coverage--reliability dynamics during supervised pretraining affect posttraining: checkpoints with similar pass@$1$ can yield different posttraining performance. Our coverage--reliability criterion selects checkpoints that outperform later checkpoints favored by mean task success and held-out action MSE (App.~\ref{app:coverage_reliability_full_results}).
\end{takeawaybox}


\begin{wrapfigure}{r}{0.45\columnwidth}
    \centering
    \vspace{-0.5em}
    \includegraphics[width=\linewidth]{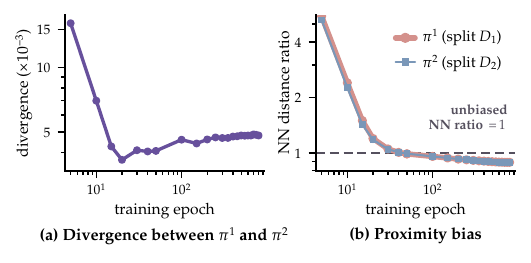}
    \vspace{-0.8cm}
    \caption{\textbf{Independent splits expose dataset-specific fitting.}
    (a) Mean-squared prediction difference between two policies trained on disjoint splits, evaluated on the same held-out inputs with matched diffusion noise; predictions first converge, then diverge.
    (b) Ratio between the average nearest-neighbor distance from policy samples to their own training split and the average nearest-neighbor distance from held-out expert actions to that same training split; decreasing values indicate increasing bias toward the training split.}
    \label{fig:split_divergence}
    \vspace{-0.5em}
\end{wrapfigure}

\textbf{Loss of pass@$k$ performance during training corresponds with overfitting on the empirical dataset, and concentrates on low noise levels.}
We next examine the learning dynamics behind with the late-checkpoint loss of coverage. We find that pretraining initially improves generalization but later increasingly fits the finite training set. In particular, later checkpoints have higher held-out denoising loss near the low-noise end despite lower training loss, with larger relative increases where observation support is sparse.

We first test the shift toward training-set overfitting. On \texttt{Threading D2}, we split 1,000 demonstrations into two disjoint 200-demo training sets, $D_1$ and $D_2$
, and a shared 600-demo held-out set, $D_3$.
We train separate $\pi_\tau^{(1)}$ and $\pi_\tau^{(2)}$ on $D_1$ and $D_2$, respectively. 
Let $\bar a_\tau^{(i)}(o,z)$ denote the action chunk generated by $\pi_\tau^{(i)}$ from observation $o$ and initial noise $z$. Fig.~\ref{fig:split_divergence}(a) measures prediction disagreement 
$D_{\mathrm{split}}(\tau)=
\mathbb{E}_{o\sim D_3,z}
[\|\bar a_\tau^{(1)}(o,z)-\bar a_\tau^{(2)}(o,z)\|_2^2]$.
Lower $D_{\mathrm{split}}$ means the two policies agree more on the same held-out inputs. This quantity initially decreases but increases again during later training, suggesting that each policy increasingly fits its own training split. This is corroborated by Fig.~\ref{fig:split_divergence}(b), which measures the ratio $R_i$ between nearest neighbor distances from actions $\bar a_{\tau}^{(i)} \sim \pi_{\tau}^{(i)}$ and $D_i$, divided by those from $a^\star \sim D_3$ (see App.~\ref{app:overfitting_diagnosis} for details). Here, $R_i<1$ means policy samples are closer to their own training split than held-out expert actions are. As pretraining proceeds, $R_i$ decreases for both policies. Taken together, these measurements support increasing fitting to the respective training splits during later training.

These split-data results are consistent with prior work showing that diffusion models shift from generalization toward training-set fitting and memorization under prolonged optimization~\citep{gu2023memorization,baptista2025memorization,ye2026provable,bertrand2026closed,bonnaire2026diffusion,Favero2025BiggerIA}. In our image-conditioned robot policies, App.~\ref{app:overfitting_diagnosis} further shows that later checkpoints have higher held-out denoising loss than earlier checkpoints near the low-noise end, despite having lower training loss at every noise level (Fig.~\ref{fig:app_noise_resolved_patching}(a,b)). The relative increase in held-out loss is also greater for observations with sparse training support and begins at higher noise (Fig.~\ref{fig:condition_support_overfit}). The intuition is that as the noise level decreases, the set of clean actions compatible with a noisy action shrinks; when observation support is sparse, fewer demonstrations also have nearby observations, leaving fewer training examples compatible with both the observation and the noisy action.

\begin{takeawaybox}
Supervised pretraining initially improves generalization, but later checkpoints increasingly fit the finite training set. Held-out denoising loss rises near the low-noise end, particularly for observations with sparse training support.
\end{takeawaybox}

\section{Diagnosing the Coverage--Reliability Tradeoff and Selecting Checkpoints Offline}
\label{sec:offline_diagnosis}

Having characterized the transition toward finite-data overfitting, we next ask how to identify the coverage--reliability balance before overfitting, without explicitly evaluating pass@$k$ performance through closed-loop policy deployment.

\begin{wrapfigure}{r}{0.45\columnwidth}
    \centering
    \vspace{-0.5em}
    \includegraphics[width=\linewidth]{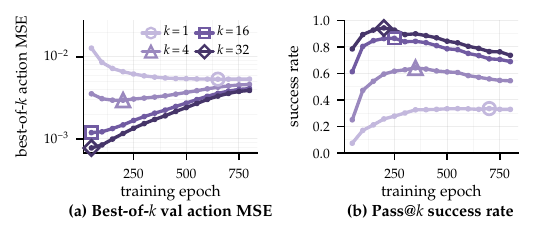}
    \vspace{-0.8cm}
    \caption{\textbf{Best-of-$k$ action MSE selects earlier checkpoints than closed-loop pass@$k$.}
    (a) Held-out best-of-$k$ action MSE across training checkpoints for different $k$.
    (b) Closed-loop pass@$k$ success rate across the same checkpoints.
    Matching colors and marker shapes denote the same value of $k$ across panels;
    hollow markers indicate the minimum MSE in (a) and maximum pass@$k$ in (b).
    As $k$ increases, both optima shift earlier, but the best-of-$k$ MSE minimum precedes the corresponding pass@$k$ peak, with a larger separation at higher $k$.
    }
    \label{fig:bok_vs_passk}
    \vspace{-0.5cm}
\end{wrapfigure}

\textbf{Best-of-$k$ action MSE captures the favorable tail but selects too early.}
Because diffusion denoising loss is often a poor indicator of closed-loop performance~\citep{allshire2026scalable}, we use held-out action MSE for offline evaluation throughout the paper~\citep{allshire2026scalable}. A tempting offline metric for approximating closed-loop pass@$k$ is then held-out best-of-$k$ action MSE. Given a held-out dataset $\mathcal{D}_{\mathrm{val}}$, we sample $k$ action chunks at each observation and retain the one closest to the expert action:
\begin{equation}
\label{eq:best_of_k_mse}
B_k(\tau)
=
\mathbb{E}_{(o,a^\star)\sim\mathcal{D}_{\mathrm{val}}}
\left[
\min_{j\le k}
\ell\!\left(A_{\tau,j}(o),a^\star\right)
\right].
\end{equation}
Fig.~\ref{fig:bok_vs_passk} shows that larger $k$ shifts the optimum of $B_k$ earlier, as it does for pass@$k$. For small $k$, the two optima remain relatively close; as $k$ increases, the best-of-$k$ minimum increasingly precedes the corresponding closed-loop pass@$k$ peak. This mismatch arises because $B_k$ measures only the favorable tail independently at each observation, whereas closed loop control requires reliable actions throughout an entire rollout trajectory. An early checkpoint can therefore attain low large-$k$-$B_k$ while still assigning substantial probability to poor action samples that destabilize execution. This motivates us to move beyond the minimum over samples and examine the full held-out action MSE distribution.

\textbf{Held-out action-MSE quantiles better reflect coverage and reliability.}
Best-of-$k$ MSE uses only the minimum error among sampled actions. Here, we now examine the full distribution of action MSE under policy samples. For held-out pair $(o_i,a_i^\star)$ and checkpoint $\tau$, draw $A_j\sim\pi_\tau(\cdot\mid o_i)$, $j=1,\ldots,M$, and let
$Q_{i,\tau}(u)$ be the $u$-quantile of $\{\ell(A_j,a_i^\star)\}_{j=1}^{M}$.
We average each quantile across held-out observations,
\begin{equation}
\label{eq:mean_error_quantile}
\bar Q_\tau(u)
=
\frac{1}{N}\sum_{i=1}^{N}Q_{i,\tau}(u).
\end{equation}

Fig.~\ref{fig:fig4_coverage_reliability}(a) shows a more systematic ordering: lower quantiles reach their minima earlier and then worsen, whereas higher quantiles continue improving later in supervised pretraining. Fig.~\ref{fig:fig4_coverage_reliability}(b) illustrates this directly: the early checkpoint can produce lucky draws closer to the expert action, but its samples are substantially more \emph{dispersed}; the late checkpoint produces more \emph{concentrated} samples, but its best sample is farther from the expert.

\newcommand{\taufin}{H}
\newcommand{\rtau}{r_{\tau \mid H}}
\newcommand{\rcov}{r_{\mathrm{cov} \mid H}}
\newcommand{\rrel}{r_{\mathrm{rel} \mid H}}

\begin{figure*}[t]
    \centering
    \includegraphics[width=\textwidth]{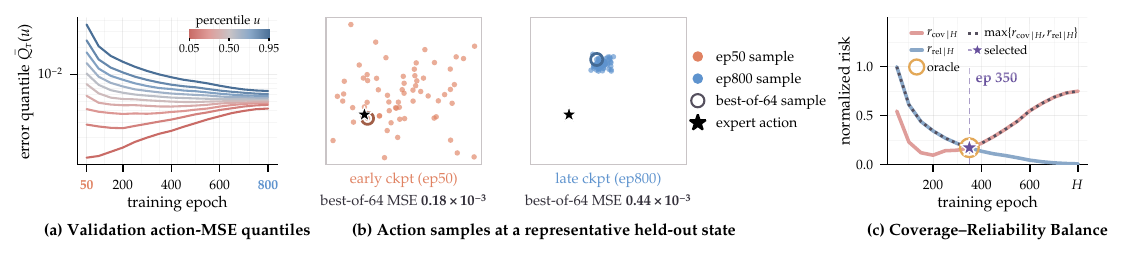}
    \vspace{-0.5cm}
    \caption{\textbf{Held-out action-MSE quantiles expose the coverage--reliability balance.}
    (a) Held-out action-MSE quantiles across training, colored by percentile $u$ from $0.05$ (red) to $0.95$ (blue).
    (b) For a held-out observation, 64 matched-noise samples from \textcolor{selectedorange}{ep50} and \textcolor{finalblue}{ep800}, plus the expert action, are projected with PCA; rings mark the minimum-MSE sample.
    \textcolor{selectedorange}{ep50} is more dispersed yet has lower best-of-64 MSE, whereas \textcolor{finalblue}{ep800} contracts but its best sample lies farther from the expert.
    Consistent with (a), lower quantiles favor earlier checkpoints, while higher quantiles improve later.
    (c) We summarize the lower and upper halves as $\rcov(\tau)$ and $\rrel(\tau)$ and select the checkpoint minimizing $\max\{\rcov(\tau),\rrel(\tau)\}$; the selected ep350 matches the closed-loop oracle.}
    \label{fig:fig4_coverage_reliability}
    \vspace{-0.4cm}
\end{figure*}

\textbf{A distributional criterion identifies the coverage--reliability balance.}
Given the previous findings, we propose to use lower quantiles as an offline measure of \emph{coverage}, i.e. whether low-MSE action chunks remain accessible under the policy, and higher quantiles as an measure of \emph{reliability}, how consistently sampled action chunks remain close to the demonstrated action. 
Based on this, we define the following normalized quantile score for training budget $H$:

\begin{equation}
\label{eq:normalized_quantile_risk}
\rtau(u)
=
\frac{
\log \bar Q_\tau(u)-\min_{s\le \taufin}\log \bar Q_s(u)
}{
\max_{\tau'\le \taufin}\log \bar Q_{\tau'}(u)-\min_{s\le \taufin}\log \bar Q_s(u)
}
\end{equation}
Here, $\rtau(u)=0$ means checkpoint $\tau$ has the lowest observed MSE at percentile $u$ among checkpoints up to $H$, while $\rtau(u)=1$ means the highest; intermediate values indicate its normalized position between these two extremes. Let $\mathcal U$ denote the discrete set of evaluated quantile levels. To evaluate policy coverage and reliability, we average lower and upper halves of the quantile profile as
\begin{equation}
\label{eq:coverage_reliability_halves}
\rcov(\tau)
=
\operatorname{mean}_{u\in\mathcal U,\;u<0.5} \rtau(u),
\qquad
\rrel(\tau)
=
\operatorname{mean}_{u\in\mathcal U,\;u\ge0.5} \rtau(u).
\end{equation}
using $u=0.5$ to split the profile evenly without additional tuned threshold. We then select
\begin{equation}
\label{eq:coverage_reliability_selector}
\tau^\star_H
=
\arg\min_{\tau\le H}
\max\left\{
\rcov(\tau),
\rrel(\tau)
\right\}.
\end{equation}
\vspace{-0.4cm}

The criterion therefore selects the checkpoint that minimizes the larger of $\rcov$ and $\rrel$. Unlike best-of-$k$, which uses only the minimum, it accounts for both lower and upper quantiles; unlike average validation MSE, it does not collapse the error distribution to a single average. By minimizing the worse of the two scores, the criterion avoids selecting a checkpoint with strong coverage but poor reliability, or vice versa. 
Lastly, because the criterion is defined relative to the quantile values, it also does not require task-specific calibration over absolute action-MSE thresholds.

Fig.~\ref{fig:fig4_coverage_reliability}(c) illustrates the criterion on the 200-demonstration \texttt{Threading D2} run. Using only offline held-out action MSE, the criterion recovers the closed-loop oracle at epoch 350. We define the closed-loop oracle by applying the same minimax balancing principle across the full pass@$k$ profile. The criterion is also robust to the choice of training horizon once the coverage--reliability dynamics has stabilized: after lower quantiles begin to worsen and higher quantiles saturate, the selected checkpoint remains relatively stable as training continues.    Extending training substantially beyond this point only marginally shifts the selected checkpoint (App.~\ref{app:coverage_reliability_full_results}). App.~\ref{app:coverage_reliability_full_results} further validates the criterion across MimicGen tasks and dataset sizes, as well as multitask pretraining on RoboCasa.


\begin{takeawaybox}
Best-of-$k$ action MSE uses only the minimum-error sample and selects checkpoints too early, especially at large $k$.
Across the full held-out action-MSE distribution, lower quantiles favor earlier checkpoints while higher quantiles improve later, enabling an offline criterion that balances coverage and reliability without closed-loop evaluation.
\end{takeawaybox}

\section{Generative Augmentation Preserves Coverage for Downstream posttraining}
\label{sec:data_aug}

\begin{figure*}[!t]
    \centering

    \includegraphics[width=\textwidth]{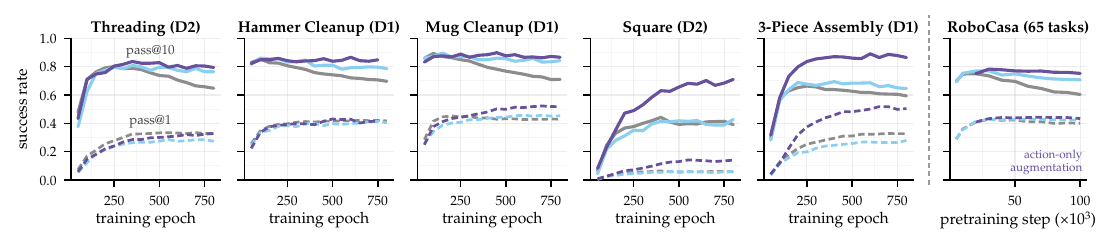}
    \vspace{-0.6cm}

    \includegraphics[width=0.50\textwidth]{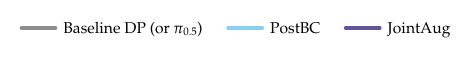}
    \vspace{-0.3cm}

    \caption{\textbf{During pretraining, generative augmentation preserves coverage while maintaining, and in some tasks improving, reliability (pass@$1$).}
    Across most MimicGen tasks and RoboCasa, \BaselineDP{} loses pass@$10$ at later checkpoints.
    \JointOA{} prevents this loss of coverage while maintaining or improving pass@$1$.
    \PostBC{} also preserves pass@$10$ better than \BaselineDP{}, but can do so at the cost of pass@$1$.}
    \label{fig:data_aug_pretraining}

    \vspace{0.2cm}

    \includegraphics[width=\textwidth]{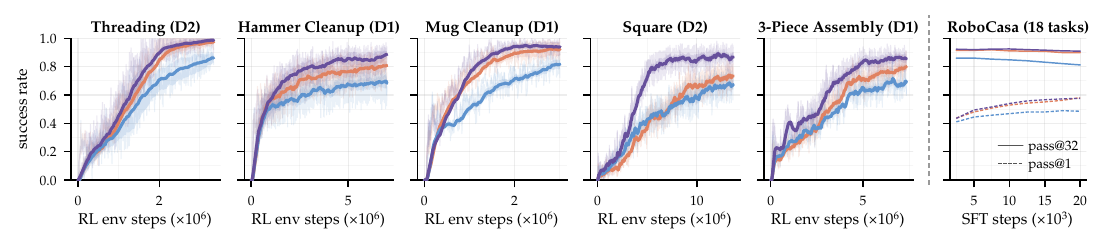}
    \vspace{-0.55cm}

    \includegraphics[width=\textwidth]{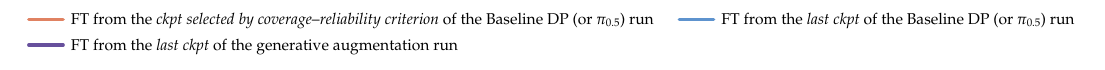}
    \vspace{-0.5cm}

    \caption{\textbf{Generative augmentation produces stronger checkpoints for posttraining.}
    Columns correspond to Fig.~\ref{fig:data_aug_pretraining}.
    We compare posttraining from the baseline checkpoint selected by our coverage--reliability criterion
    (Sec.~\ref{sec:offline_diagnosis}), the final \BaselineDP{} checkpoint, and the final checkpoint from our generative augmentation run.
    Across MimicGen tasks, the final generative-augmentation checkpoint substantially outperforms the final baseline and often matches or exceeds the selected checkpoint.
    On RoboCasa, it matches the selected checkpoint on both pass@$1$ and pass@$32$.}
    \label{fig:data_aug_finetune}
    \vspace{-0.3cm}
\end{figure*}


Although our proposed coverage--reliability criterion (Sec.~\ref{sec:offline_diagnosis}) selects a favorable compromise without closed-loop evaluation, it is constrained to the observed coverage--reliability Pareto frontier of that run. 
This raises a more fundamental question: is this tradeoff unavoidable? Can a training intervention give us the best of both worlds by preserving strong large-$k$ pass@$k$ while continuing to improve reliability? If so, we can raise the coverage--reliability Pareto frontier. 

Sec.~\ref{sec:passk_dynamics} links the loss of coverage to overfitting to the finite demonstration data during policy training. In App.~\ref{app:overfitting_diagnosis}, we show that the held-out denoising score-matching loss increases more sharply late in training for the observation-conditioned model than for its unconditional counterpart. Indeed, \citet{gu2023memorization} show that even conditioning diffusion models on uninformative random labels can substantially exacerbate memorization. This suggests a direct intervention: expand supervision beyond the observed $(o,a)$ pairs using fresh synthetic observation--action pairs that reflect the population-level expert distribution.

\textbf{Joint $(o,a)$ augmentation.}
To expand supervision beyond observed $(o,a)$ pairs while preserving observation--action consistency, we propose \emph{joint $(o,a)$-augmentation} (\JointOA{}), which trains a generative model $G_\gamma(o,a)$ with checkpoints indexed by $\gamma$ on the joint demonstration distribution and uses generated pairs for policy training. We save generator checkpoints at fixed intervals, and aim to identify a sampling window after the generator has learned population-level structure but before substantial overfitting. To this we end record each checkpoint's ratio between generated actions' nearest-neighbor distances to training expert actions and held-out expert actions (App.~\ref{app:impl_jointaug}). A ratio of $1$ indicates equal proximity to the two sets, and values below one indicate that generated samples are closer to the training data. We let $\gamma_0$ be the first checkpoint where the ratio falls below one, when generated samples first become closer to training than held-out expert actions. We then sample synthetic $(\tilde o,\tilde a)$ pairs uniformly from $\gamma_0$ and the four subsequent checkpoints spaced 100 epochs apart. Because checkpoints are saved relatively densely over this stage of generator training, the five-checkpoint window spans only the transition from population-level generalization toward early fitting of the empirical training set. We use five checkpoints as a simple default heuristic across tasks rather than tuning the window for each task. App.~\ref{app:jointaug_robust_to_ckpt} shows that performance is robust to the exact generator checkpoint selection.


Then policy training uses equal amounts of real and generated supervision:
\begin{equation}
\label{eq:gen_aug_training_objective}
\mathcal L_{\mathrm{aug}}
=
\tfrac12\,
\mathbb E_{(o,a)\sim\mathcal D}
[\mathcal \Ldp(o,a)]
+
\tfrac12\,
\mathbb E_{(\tilde o,\tilde a)\sim G_{\mathrm{mix}}}
[\mathcal \Ldp(\tilde o,\tilde a)],
\end{equation}
where $G_{\mathrm{mix}}$ is the uniform mixture of the five selected generator
checkpoints. Real samples retain the clean expert targets; generated pairs expand supervision beyond the finite data. Because observations and actions are generated jointly, their correspondence is naturally preserved.

To evaluate \JointOA{}, we compare against baseline Diffusion Policy and posterior behavior cloning (\PostBC{}, \cite{wagenmaker2025posterior}) across the full training horizon. \PostBC{} aims to improve policy coverage by estimating observation-conditioned action covariance with a bootstrapped ensemble and perturbing action targets $\tilde a$ at clean observations $o$. In that regime, the perturbed action is not constrained to remain compatible with the clean observation. In contrast, \JointOA{} ensures $(\tilde o, \tilde a)$ is learned from the correct joint distribution (and thus, $\tilde a \mid \tilde o$ is consistent).

In the first five panels of Fig.~\ref{fig:data_aug_pretraining}, we compare the methods across MimicGen tasks. \JointOA{} achieves higher pass@$10$ than both \PostBC{} and the baseline DP run and prevents the baseline's late-training loss of coverage. Its pass@$1$ matches the baseline and sometimes improves it. \PostBC{} also improves pass@$10$, but on some tasks lowers pass@$1$, likely because its perturbed actions can be incompatible with the clean observations. Moreover, unlike \PostBC, which perturbs only action targets, \JointOA generates new observation–action pairs to further mitigate the conditional overfitting observed in App.~\ref{app:overfitting_diagnosis}. App.~\ref{app:jointaug_action_cloud_comp} analyzes the sampled action distributions. \JointOA{} preserves the favorable low-error tail and reduces typical-sample error, helping explain its gains in both coverage and reliability.
\JointOA{} also provides checkpoints with stronger residual-RL initializations than the final baseline checkpoints (Fig.~\ref{fig:data_aug_finetune}), often matching or exceeding the checkpoint selected by our coverage--reliability criterion from baseline run. By preventing the late-training loss of coverage, \JointOA{} also reduces the need for post hoc checkpoint selection. We further find that its performance is robust to the choice of generator checkpoint; more ablations and baseline comparisons are in App.~\ref{app:jointaug_supp}.

\textbf{A scalable action-only variant.}
For large-scale VLA pretraining, joint generation is expensive, so on RoboCasa we use a lightweight action-only variant, similar in spirit to \PostBC{} but using the policy's own action variation rather than a bootstrapped ensemble for covariance estimation, thereby avoiding the need to tune the perturbation scale. After a standard pretraining run, we select $c^\star$ using our coverage--reliability criterion and estimate action covariance $\Sigma_{c^\star}(o)$ from multiple policy samples at each training observation. We then resume pretraining from $c^\star$ with perturbed targets
$\tilde a=a+\epsilon$, where
$\epsilon\sim\mathcal N(0,\Sigma_{c^\star}(o))$.
This injects perturbations calibrated to the action spread at $c^\star$ without shifting each training target in expectation, $\mathbb E_{\epsilon}[\tilde a\mid a,o]=a$. Fig.~\ref{fig:data_aug_pretraining} and~\ref{fig:data_aug_finetune} show that the variant preserves large-$k$ pass@$k$ as pass@$1$ improves, and achieves SFT performance comparable to the selected early checkpoint.

\begin{takeawaybox}
Generative augmentation can preserve coverage while allowing reliability to improve, producing late-pretraining checkpoints with stronger pass@$k$ profiles and better downstream posttraining than standard final checkpoints.
\end{takeawaybox}

\section{Coverage--Reliability Dynamics On a Real Robot}
\label{sec:real_robot}

\begin{figure*}[t]
    \centering
    \includegraphics[width=\textwidth]{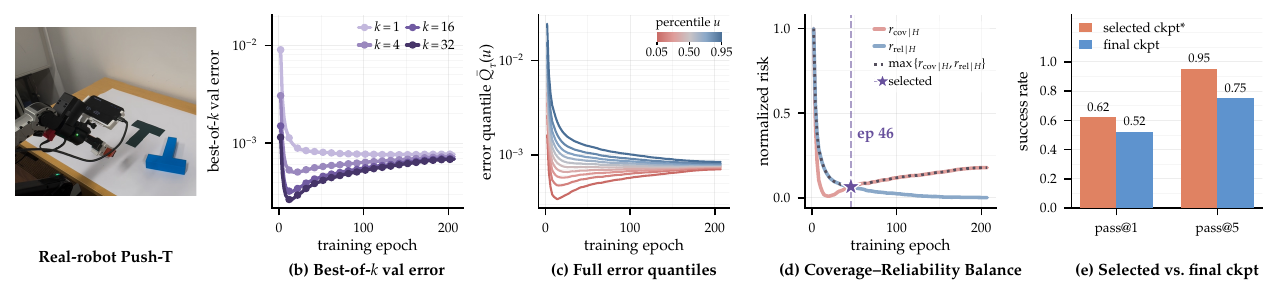}
    \vspace{-0.5cm}
    \caption{\textbf{Real-robot Push-T.}
    (a) Task setup.
    (b) Best-of-$k$ validation action MSE across training epochs.
    (c) Quantiles of validation action error across training epochs.
    (d) Normalized coverage and reliability scores with the selected checkpoint at epoch 46.
    (e) Pass@$1$ and pass@$5$ success rates of the selected (epoch 46) and final (epoch 200) checkpoints over 20 fixed initial states.}
    \label{fig:real_robot}
    \vspace{-0.4cm}
\end{figure*}

To test whether the coverage--reliability dynamics also arise in policies trained on real-robot data, we study a challenging variant of \texttt{Push-T}. We use only a wrist-mounted camera for visual observations and allow full 6-DoF end-effector control. The end-effector manipulates the T-block only through non-prehensile contact. We collect 740 human demonstrations, using 680 for training and 60 for held-out validation, and train a Diffusion Policy for 200 epochs with an action chunk length of 16. Experiment details are in App.~\ref{app:real_robot_impl}. The best-of-$k$ validation action MSE and error quantiles exhibit the same coverage--reliability tradeoff observed in simulation (Fig.~\ref{fig:real_robot}(b,c)). Applying our coverage--reliability balance criterion (Sec.~\ref{sec:offline_diagnosis}) selects the checkpoint at epoch 46 (Fig.~\ref{fig:real_robot}(d)). We compare the selected checkpoint against the final checkpoint at epoch 200 using five rollouts from each of 20 fixed initial states. The selected checkpoint outperforms the final checkpoint by 10\% on pass@$1$ and 20\% on pass@$5$ (Fig.~\ref{fig:real_robot}(e)).

\begin{wraptable}{r}{0.47\columnwidth}
    \centering
    \vspace{-0.6em}
    \footnotesize

    \begin{tabular*}{\linewidth}{@{\extracolsep{\fill}}lcc@{}}
        \toprule
        \textbf{Rollouts} & \textbf{PSNR}$\uparrow$ & \textbf{LPIPS}$\downarrow$ \\
        \midrule
        \multicolumn{3}{@{}l}{\emph{Chunk 1}} \\
        Selected & \textbf{27.9} & \textbf{0.112} \\
        Final    & 26.8          & 0.120 \\
        \midrule
        \multicolumn{3}{@{}l}{\emph{Chunk 2}} \\
        Selected & \textbf{25.2} & \textbf{0.136} \\
        Final    & 24.1          & 0.145 \\
        \bottomrule
    \end{tabular*}
    \caption{\textbf{World-model prediction quality.}
    Models are trained on demonstrations and rollouts from the selected or final checkpoint.
    Chunk 1 is predicted from two ground-truth context frames; Chunk 2 is predicted
    autoregressively from the model's Chunk-1 output.}
    \label{tab:real_robot_wm}

    \vspace{-0.5em}
\end{wraptable}

To investigate whether differences in policy coverage affect downstream model learning, we train two action-conditioned world models on the same demonstration data, augmented with 200 rollout trajectories from either the selected early checkpoint or the final checkpoint. We include all rollouts regardless of task success for world model training. For held-out validation, we collect another 100 rollout trajectories from intermediate checkpoints, with 20 trajectories each from epochs 75, 100, 125, 150, and 175.

The world model uses a frozen Stable Diffusion VAE to encode each image frame into a latent. We train a spatio-temporal DiT from scratch and inject action conditioning additively at each frame. As in policy training, we use an action chunk length of 16. Given two history frames and a 16-step action chunk, the model predicts the next four frames. At evaluation, we first predict four frames from two ground-truth history frames and the corresponding action chunk. We then predict the following four frames autoregressively from the model's own predictions and the next action chunk. We average PSNR and LPIPS over the four predicted frames in each chunk. As shown in Tab.~\ref{tab:real_robot_wm}, the world model trained with selected-checkpoint rollouts achieves better prediction quality than the model trained with final-checkpoint rollouts. This result suggests that preserving policy coverage can benefit downstream world-model learning. Additional visualizations of the world model's prediction are in App.~\ref{app:real_robot_impl}.

\vspace{-0.3cm}
\section{Abridged Related Work}
\vspace{-0.3cm}
Due to space constraints, we discuss key related work here and defer more comprehensive review to Appendix~\ref{sec:related_work}. 
In language models, coverage of high-quality outputs has emerged as an important property for efficient posttraining and test-time scaling~\citep{foster2025good,chen2026coverage}. Modern generative robot policies \citep{chi2025diffusion,kim2024openvla,black2024pi_0} are likewise followed by supervised finetuning, RL, or inference-time steering \citep{kim2025fine,intelligence2025pi,intelligence2026pi}, yet few works study how properties acquired during pretraining shape these later stages. PostBC and TMRL explicitly modify pretraining to improve subsequent RL by preserving action coverage or exploration \citep{wagenmaker2025posterior,hong2026tmrl}. We instead study how coverage and reliability change during standard supervised pretraining itself and how they affect posttraining. Mirroring work showing that diffusion models can transition from generalization toward finite-sample fitting over training \citep{bonnaire2026diffusion,Favero2025BiggerIA}, we characterize this learning dynamic in robot policies and connect it to closed-loop coverage and posttraining.

\vspace{-0.2cm}
\section{Conclusion and Future Work}
\vspace{-0.2cm}
This work studies how coverage--reliability dynamics during supervised robot-policy pretraining affect posttraining performance, and develops practical methods for offline checkpoint selection and coverage-preserving pretraining. An important direction for future work is to jointly develop principled data collection strategies and improved pretraining methods: demonstration datasets should have both broad coverage of task conditions and sufficient local density, and training objectives, architecture designs, and optimization strategies should improve reliability without sacrificing coverage or generalization. Another future direction is to develop a deeper theoretical understanding of how policy distributions evolve during pretraining and how these changes affect subsequent posttraining, which could guide both data collection and policy optimization. More broadly, our findings motivate a unified view of robot-policy pretraining and posttraining, in which pretrained policies are evaluated not only by their immediate performance but also by how effectively they can be improved through posttraining.

\vspace{-0.2cm}
\section*{Acknowledgements}
\vspace{-0.2cm}
The authors thank all members of REALab and Huy Ha for their feedback on an earlier draft of this manuscript, and Jack Joler for data collection for the real-robot experiments. ZS is supported by the Stanford Graduate Fellowship. This work was supported in part by the NSF Award \#2143601, \#2037101, and \#2132519. The views and conclusions contained herein are those of the authors and should not be interpreted as necessarily representing the official policies, either expressed or implied, of the sponsors.


\vspace{-0.2cm}
\subsection*{AI Use Statement}
\vspace{-0.2cm}

We used generative AI tools to polish the writing, search for relevant literature, and help with parts of the code implementation. We did not use them to generate data, derive mathematical results, develop research ideas, or interpret experimental results. The authors checked all AI-assisted work: we tested the code, verified every cited paper against its source, and confirmed the text accurately reflects our contributions. We take full responsibility for the content of this paper.


\bibliography{iclr2027_conference}
\bibliographystyle{iclr2027_conference}

\newpage
\appendix
\appendix

\startcontents[appendix]

\section*{Appendix Contents}
\begingroup
\linespread{0.97}\selectfont
\printcontents[appendix]{}{1}{\setcounter{tocdepth}{2}}
\endgroup

\clearpage

\section{Related Work}
\label{sec:related_work}

\appsubsec{Pretraining--posttraining Interactions in Robot Learning}
\label{app:related_work_pretraining_posttraining}
Language-model research provides strong motivation for studying pretraining and posttraining jointly. Recent work shows that performance within one stage in isolation can poorly predict the outcomes after the next. Models with similar pretraining validation loss can differ substantially in thei finetuning performance~\citep{liu2023same}, and prolonged pretraining can reduce subsequent finetuning performance~\citep{springer2025overtrained}. In SFT--RL pipeline, higher SFT performance does not always predict stronger post-RL performance; stronger SFT checkpoints can even perform worse after the same RL procedure~\citep{kang2026quagmires,zhang2026good}. Recent theory instead highlights coverage, the probability mass assigned to high-quality outputs, as an important property for test-time sampling~\citep{foster2025good,chen2026coverage}. Empirically, large-$k$ pass@$k$ can predict post-RL performance better than SFT task performance~\citep{kang2026quagmires}. TailSFT similarly improves posttraining by modifying SFT to preserve broader output coverage~\citep{malladi2026tailsft}.
Robot learning increasingly follows a similar staged pipeline: BC or pretraining \citep{chi2025diffusion,kim2024openvla,black2024pi_0} followed by SFT~\citep{kim2025fine,bjorck2025gr00t,nasiriany2026robocasa365}, RL~\citep{intelligence2025pi,wagenmaker2025steering}, or inference-time steering~\citep{intelligence2026pi,wang2025inference,chen2026steerable}. Yet few works study how properties learned during pretraining affect posttraining. PostBC proves that standard BC can fail to guarantee coverage of the demonstrator's actions, which is necessary for RL finetuning to reach the demonstrator's performance. It instead models the posterior over demonstrator action to guarantee such coverage and improve RL finetuning~\citep{wagenmaker2025posterior}. TMRL broadens coverage during pretraining by adding noise to the policy context, then lets RL control the noise level during finetuning~\citep{hong2026tmrl}. Both methods redesign pretraining with noise injection specifically for downstream RL finetunig. We instead study how a policy's finetunability changes over the course of standard supervised pretraining. We identify an evolving coverage--reliability tradeoff, show that it affects multiple forms of posttraining, and use this to design offline checkpoint selection criterion and training intervention.

\appsubsec{Generalization and Memorization in Diffusion Models}
\label{app:related_work_dm_learning_dynamics}
Finite-data diffusion objectives admit closed form solutions that reproduce the training set, and several works characterize how model capacity, dataset size, conditioning, and optimization determine whether trained models approach these memorizing solutions~\citep{gu2023memorization,baptista2025memorization,ye2026provable,biroli2024dynamical}. Recent papers further show a temporal separation in training dynamics: models first learn population-level structure that generalizes beyond the training set, while prolonged optimization can increasingly fit individual training examples~\citep{bonnaire2026diffusion,Favero2025BiggerIA}. Related work on flow matching likewise studies why learned models generalize despite an empirical optimum that memorizes the data~\citep{bertrand2026closed}. These findings have motivated interventions including early stopping, limiting model capacity~\citep{baptista2025memorization,ye2026provable,Favero2025BiggerIA}, training from corrupted examples~\citep{daras2023ambient,shah2025does}, and perturbing conditioning input~\citep{somepalli2023understanding}. Most of this literature studies unconditional, class-conditioned, or text-conditioned generation of independent samples. Robot policies instead condition action generation on dense, high-dimensional observations and repeatedly sample the policy for multi-turn closed-loop interaction. \citet{he2026demystifying} adopt a static view of trained diffusion policies and argue that, on small robot datasets, they largely retrieve actions from nearby training examples. We instead study how the policy distribution dynamically evolves over optimization, how this evolution changes closed-loop reliability and coverage and posttraining.

\appsubsec{Robot Policy Evaluation and Model Selection}
\label{app:related_work_policy_eval}
Closed-loop rollouts remain the standard for evaluating robot policies, but reliable comparisons require many trials and careful statistical treatment~\citep{kress2024robot,barreiros2026careful,snyder2025your}. This cost motivates offline metrics for policy development and checkpoint selection. \citet{mandlekar2021matters} showed that validation loss can select substantially worse BC checkpoints than rollout evaluation. More recently, ABC found that diffusion validation loss was poorly correlated with real-world performance, whereas validation action error correlates more strongly with it~\citep{allshire2026scalable}. Other work proposes manifold adherence~\citep{pan2026much} or task-critical action error~\citep{huang2026critical} as more informative offline signals. Our setting differs because stochastic generative policies can have similar average error while differing substantially across repeated sampling. We therefore evaluate the full distribution of held-out action errors and select checkpoints by balancing coverage and reliability with a special focus for on their use in downstream posttraining.

\appsubsec{Noise Injection in Robot Policies}
\label{app:related_work_noise_injection}
Noise injection and local data augmentation methods have long been used to reduce covariate shift in behavior cloning. DART injects action noise during expert collection so demonstrations cover states reached after action perturbations~\citep{laskey2017dart}. Other work perturbs existing demonstrations and synthesizes corrective labels~\citep{ke2021grasping}, uses noise augmentation to improve closed-loop stability~\citep{block2023provable}, or generates corrective data under local continuity assumptions~\citep{ke2024ccil}. Recent theory shows that expert action noise can mitigate exponential compounding error in continuous-control imitation learning~\citep{zhang2025action}. PostBC~\citep{wagenmaker2025posterior} and TMRL~\citep{hong2026tmrl} similarly inject noise into actions or observation conditioning to broaden coverage for downstream RL. Our joint (o,a)-augmentation instead learns the joint observation--action distribution from the demonstrations and generates new $(o,a)$ pairs together. Thus, the dataset itself determines both the observation variation and its compatible action, without additional expert collection, a hand-designed external noise distribution, or a local analytical dynamics model.

\clearpage
\section{Success Polarization Across Evaluation Conditions}
\label{app:condition_redistribution}

Sec.~\ref{sec:passk_dynamics} showed that pass@$k$ evolves asynchronously during supervised pretraining, and attributed this behavior to a redistribution of success across initial evaluation conditions. Here, we examine this redistribution directly at the condition level and make explicit why it causes larger-$k$ pass@$k$ performance to decline first.

For the 200-demonstration Threading-D2 experiment, we evaluate every checkpoint on the same 500 evaluation conditions, with 32 rollout attempts per condition (pass@32). For each condition, we count the number of successful rollouts and track how the distribution of these counts changes over the course of supervised training.

\begin{wrapfigure}{r}{0.45\columnwidth}
    \centering
    \vspace{-0.8em}
    \includegraphics[width=\linewidth]
    {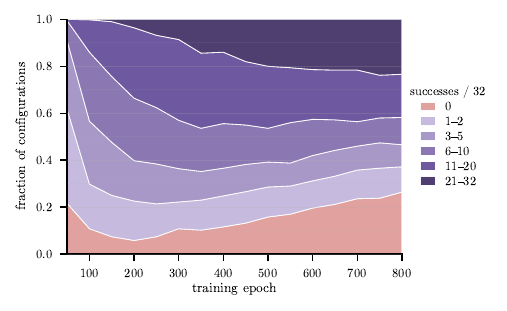}
    \caption{\textbf{Distribution of condition-level success counts over supervised pretraining.}
    For each checkpoint, we run pass@32 on each of 500 evaluation conditions and bin each condition by its number of successful rollouts within 32 trials. The stacked bands show the fraction of conditions in each success-count bin; each vertical slice sums to one. During late training, the fraction of conditions with zero successes increases, while the fraction with high success counts also increases.}

    \label{fig:app_success_bins}
    \vspace{-0.8em}
\end{wrapfigure}

\paragraph{Late training polarizes success across conditions.}
Fig.~\ref{fig:app_success_bins} makes the redistribution described in Sec.~\ref{sec:passk_dynamics} explicit. Early in training, the improvement is largely uniform: the fraction of conditions with zero successes decreases and more conditions achieve higher success counts. Later, however, this trend reverses for the hardest conditions. Later in training, both the fraction of conditions with zero successes and the fraction with high success counts increase. Therefore, late training \emph{polarizes} condition-wise success, improving reliability on a subset of conditions while reducing coverage on others.

\paragraph{Why polarization produces asynchronous pass@$k$ profiles.}
Let $q_\tau(c)$ denote the one-rollout success probability for condition $c$ at
checkpoint $\tau$. Then
\[
P_k(\tau)
=
\mathbb{E}_{c}
\left[
1-\bigl(1-q_\tau(c)\bigr)^k
\right].
\]
Pass@1 averages $q_\tau(c)$ linearly. For $k>1$,
$f_k(q)=1-(1-q)^k$ is concave. Thus, shifting
success probability from a lower-$q$ condition to a higher-$q$ condition while holding mean success fixed leaves
pass@1 unchanged but lowers pass@$k$.

\[
\frac{\partial P_k}{\partial \tau}
=
\mathbb{E}_{c}
\left[
k(1-q_\tau(c))^{k-1}
\frac{\partial q_\tau(c)}{\partial \tau}
\right].
\]
As $k$ increases, the factor $(1-q_\tau(c))^{k-1}$ suppresses changes on
high-success conditions more strongly. Large-$k$ pass@$k$ therefore depends
more on whether low-success conditions remain solvable. Once those conditions
begin to degrade, large-$k$ pass@$k$ begins to decrease, even if small-$k$ pass@$k$ continue to improve from improvements on already-reliable conditions.
\section{Additional Results on RL Finetuning Across the Coverage--Reliability Tradeoff}

\appsubsec{Additional Results on Residual RL}
\label{app:residual_rl_coverage_reliability}

To examine more broadly how pretraining coverage and reliability affect residual
RL finetuning (Sec.~\ref{sec:passk_dynamics}), we additionally finetune a substantially earlier checkpoint that retains more coverage but has lower pass@$1$ than the checkpoint selected by our criterion (Sec.~\ref{sec:offline_diagnosis}).

\begin{figure*}[t]
    \centering
    \includegraphics[width=\textwidth]
    {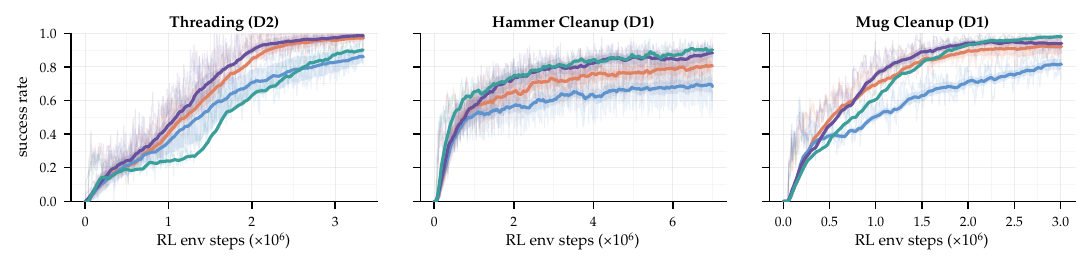}
    \vspace{-0.3em}

    \includegraphics[width=\textwidth]
    {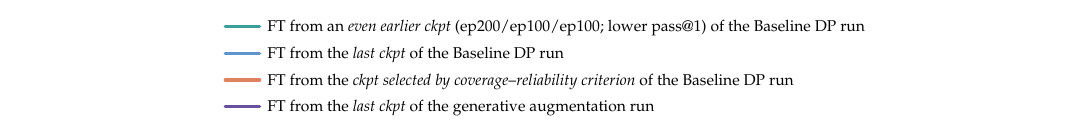}
    \vspace{-0.5em}

    \caption{\textbf{Residual RL finetuning across the coverage--reliability tradeoff.}
    We additionally finetune an earlier baseline checkpoint
    (ep.200/100/100 for \texttt{Threading}/\texttt{Hammer Cleanup}/\texttt{Mug Cleanup}) with lower pass@$1$ and higher coverage than the checkpoint selected by our criterion.
    Its performance relative to later checkpoints depends on the available online budget.}
    \label{fig:app_early_ckpt_finetuning}
\end{figure*}

\vspace{-1em}
\noindent\begin{minipage}{\textwidth}
\centering
    \includegraphics[width=0.96\linewidth]{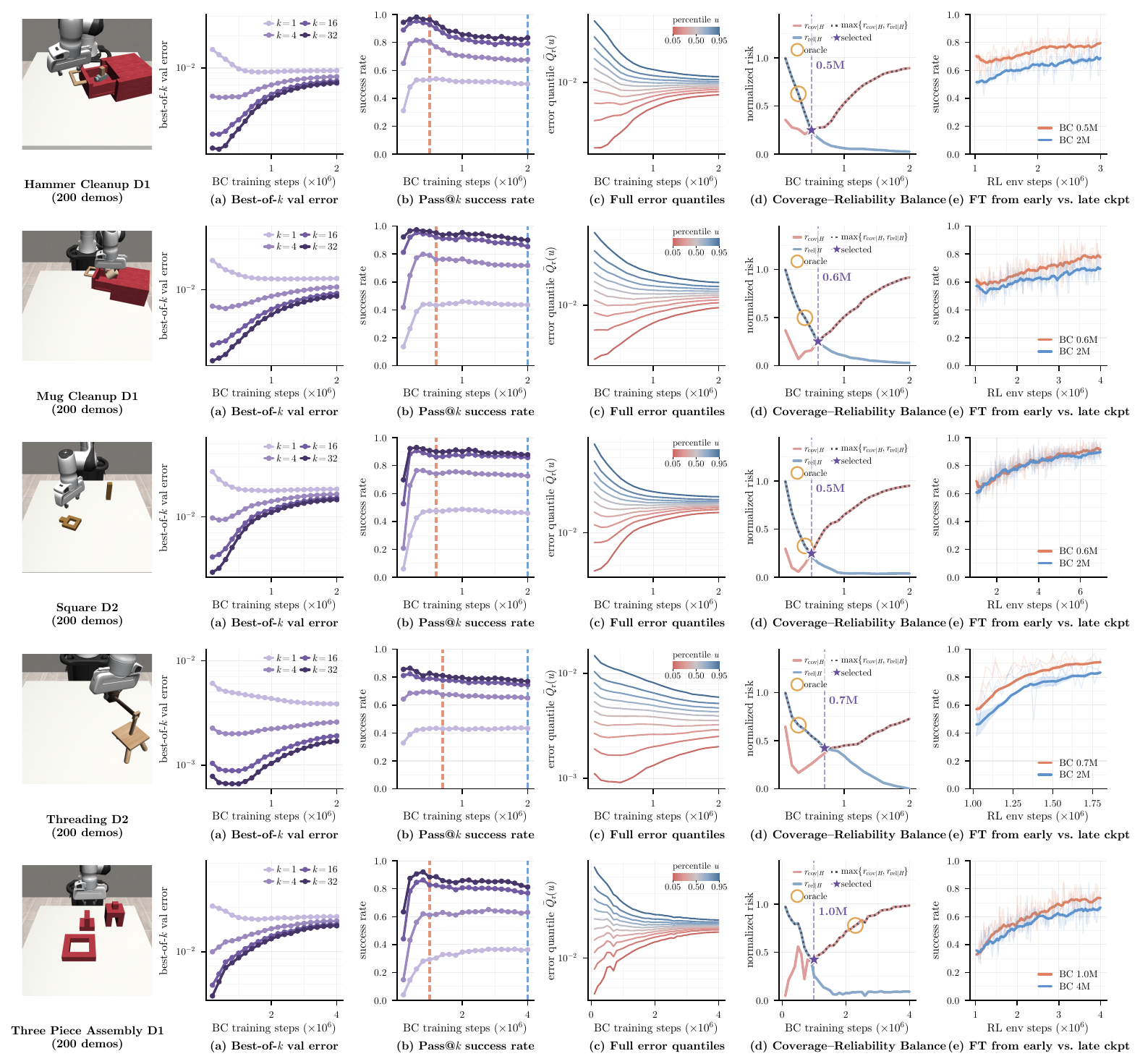}
    \vspace{-0.3em}
    \captionof{figure}{
    \textbf{Flow-matching policy pretraining and OGPO finetuning.}
    Each row shows a MimicGen task with 200 demonstrations.
    (a) Best-of-$k$ validation action MSE.
    (b) pass@$k$ during BC pretraining.
    (c) Validation error quantiles.
    (d) Coverage--reliability scores and selected checkpoint.
    (e) OGPO finetuning from the selected versus final BC checkpoint. The selected initialization generally achieves better downstream RL performance.
    }
    \label{fig:app_ogpo}
\end{minipage}

Fig.~\ref{fig:app_early_ckpt_finetuning} shows that the relative advantage of this much earlier checkpoint depends on the available online budget. Its lower pass@$1$ can hurt early residual RL performance, while its greater coverage can become beneficial with more online budget; the exact tradeoff can be task dependent. While we do not claim that our criterion selects \emph{the} optimal checkpoint for any online RL budget, it instead selects a checkpoint with substantially higher large-$k$ pass@$k$ than the final checkpoint while maintaining comparable pass@$1$; and across the evaluated budgets, this checkpoint consistently finetunes better than the final checkpoint from the same supervised run.

The exact ordering of checkpoints after RL finetuning can depend on both the RL algorithm and the task dynamics. Different RL methods explore and use online experience differently, and task dynamics affect how quickly differences between pretrained policies are amplified or corrected during closed-loop interaction. Our experiments establish the online rollout budget-dependent ordering of checkpoints for DICE-RL; we believe similar effects could also arise in other residual- or steering-based RL methods. Large-$k$ pass@$k$ should therefore be interpreted as an offline measure of how broadly a pretrained policy succeeds before finetuning, rather than as a universal predictor of checkpoint ordering after arbitrary RL finetuning procedures.

\appsubsec{Full-policy Finetuning}
\label{app:ogpo}
While DICE-RL directly depends on the quality of actions sampled from the pretrained policy, we further investigate whether checkpoint selection matters when RL can update the entire action generator. We adopt Off-Policy Generative Policy Optimization (OGPO)~\citep{patil2026ogpo} and use its official implementation. We first pretrain flow-matching policies with behavior cloning (BC). During RL finetuning, OGPO learns action-value critics through off-policy TD learning and updates the full generative policy using a PPO-style objective over sampled action-generation trajectories. All experiments use low-dimensional state observations and 200 demonstrations per task. We increase the critic warmup budget to $10^6$ steps, since shorter warmup leads to substantially slower or stalled learning on MimicGen tasks. One possible explanation is that these tasks make accurate value estimation particularly challenging, although we do not fully isolate the cause of this sensitivity here.

Figure~\ref{fig:app_ogpo}(a,b) shows that flow-matching BC policies exhibit a similar coverage--reliability trade-off: best-of-$k$ validation action MSE increases for large $k$ and large-$k$ pass@$k$ declines later in training. Notably, this observation occurs even with a relatively small five-layer MLP action network containing approximately only 1.17M parameters, suggesting that the phenomenon is not limited to policies with high-capacity networks. Figure~\ref{fig:app_ogpo}(c,d) shows the evolution of validation error quantiles and the coverage--reliability criterion used for checkpoint selection.

Figure~\ref{fig:app_ogpo}(e) compares OGPO finetuning from the selected and final BC checkpoints. The selected checkpoint achieves better performance on four of the five tasks. This result suggests that the effects of policy coverage and checkpoint selection extend beyond RL methods that directly rely on sampling and steering: even when the entire action generator is finetuned, the selected checkpoint can provide a more favorable initialization for subsequent RL. For OGPO in particular, a possible explanation is that it computes policy updates from action-generation trajectories sampled under the current policy. An initialization that places greater probability mass on useful actions may therefore generate more informative advantage-weighted updates, whereas actions that are rarely sampled initially may be difficult to reinforce under a finite interaction budget. 
\section{Diagnosing Overfitting in Diffusion Policies}
\label{app:overfitting_diagnosis}

\newcommand{\passk}{pass@$k$}
\definecolor{learnedred}{RGB}{174,55,62}
\definecolor{expertblue}{RGB}{48,93,154}

The main text establishes that extended supervised pretraining can improve pass@$1$ (reliability) while reducing large-$k$ pass@$k$ (coverage). Here, we examine how diffusion policies fit finite demonstrations over optimization time and across diffusion noise levels. We find that training denoising loss continues to decrease across noise levels, while held-out loss increases at later checkpoints toward the low-noise end. This deterioration is more pronounced in regions with sparse observation support. Furthermore, comparison with action-only diffusion trained on the same data shows that observation conditioning leads to greater late-training deterioration in held-out denoising loss, and such deterioration begins at higher diffusion noise levels. The final two low-noise denoising calls also appears to account for most of the difference in closed-loop pass@$k$ performance between early and late checkpoints.

\begin{figure*}[t]
    \centering
    \includegraphics[width=\linewidth]{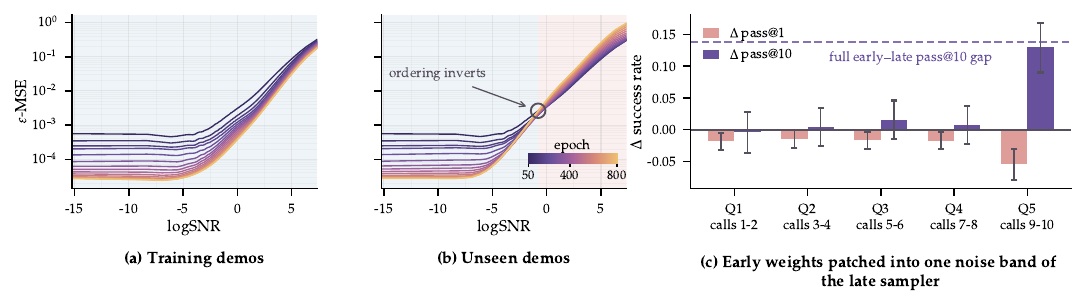}
    \vspace{-0.5em}
    \caption{\textbf{Later checkpoints exhibit higher held-out denoising loss near the low-noise end, and changes to the final denoising calls substantially affect pass@$k$ profle.}
    (a) On training demonstrations, $\epsilon$-prediction MSE decreases with further training across all noise levels.
    (b) On unseen demonstrations, later checkpoints have higher denoising loss near the low-noise end than earlier checkpoints despite lower training loss.
    (c) In a 10-step DDIM sampler, replacing any one of the first four two-step bands (Q1–Q4) of the epoch-800 checkpoint with the corresponding epoch-200 weights has little effect on pass@$1$ or pass@$10$. Replacing Q5, the final two calls, restores most of the epoch-200 checkpoint's pass@$10$ advantage while reducing pass@$1$.}
    \label{fig:app_noise_resolved_patching}
\end{figure*}

\paragraph{Held-out denoising deteriorates near the low-noise end.}
For checkpoint $\tau$ and diffusion noise level $t$, define the $\epsilon$-prediction MSE as
\begin{equation}
    L_{\epsilon,\tau}(t)
    =
    \mathbb{E}_{(o,a_0),\,\epsilon}
    \left[
        \left\|\epsilon-\epsilon_\tau(a_t,o,t)\right\|_2^2
    \right],
\end{equation}
where $a_t=\alpha_t a_0+\sigma_t\epsilon$, $\epsilon\sim\mathcal{N}(0,I)$, and the expectation over $(o,a_0)$ is evaluated separately on training and unseen demonstrations. Fig.~\ref{fig:app_noise_resolved_patching}(a,b) show this loss across diffusion noise levels for checkpoints at different training epochs, evaluated on training and unseen demonstrations, respectively. On training demonstrations, later checkpoints have lower MSE than earlier checkpoints at every noise level (Fig.~\ref{fig:app_noise_resolved_patching}(a)). On unseen demonstrations, however, this ordering reverses at the low noise end: later checkpoints have higher MSE than earlier ones at the same noise levels (Fig.~\ref{fig:app_noise_resolved_patching}(b)). This train--held-out divergence indicates \emph{overfitting} at the low noise end of the diffusion process. Since $L_{\epsilon,\tau}(t)=\mathrm{SNR}(t)L_{a_0,\tau}(t)$, expressing the error in clean-action MSE units preserves the checkpoint ordering at each fixed noise level, although it changes the absolute loss magnitudes across noise levels.

\paragraph{Independent data splits show fitting to the training split.}
Here we explain the data split experiment in main text Sec~\ref{sec:passk_dynamics} in full details. We train two policies, $\pi_\tau^{(1)}$ and $\pi_\tau^{(2)}$, on disjoint demonstration sets $D_1$ and $D_2$, and evaluate both on the same held-out set $D_3$ with matched initial diffusion noise. Their predictions on $D_3$ first become more similar, then diverge as training continues. To test whether this late divergence reflects each policy moving toward its own training split, define the nearest-neighbor action distance
\begin{equation}
    d_i(\bar a)=\min_{(o',\bar a')\in D_i}\|\bar a-\bar a'\|_2
    \label{eq:split_nn_distance}
\end{equation}
and the normalized ratio
\begin{equation}
R_i(\tau)=
\frac{
\E_{o\sim D_3}
{\color{learnedred}\E_z\!\left[d_i\!\left(\bar a_\tau^{(i)}(o,z)\right)\right]}}
{
{\color{expertblue}\E_{(o,a^\star)\sim D_3}\!\left[d_i(a^\star)\right]}}
.
\label{eq:split_bias_ratio}
\end{equation}

Here $\bar a_\tau^{(i)}(o,z)$ is an action chunk sampled from $\pi_\tau^{(i)}$ given held-out observation $o$ and initial diffusion noise $z$, and $\bar a'$ denotes the demonstrated action chunk of a training pair in $D_i$. The \textcolor{learnedred}{numerator} measures its distance to the policy's training split $D_i$; the \textcolor{expertblue}{denominator} measures the corresponding distance for held-out expert actions. Thus $R_i(\tau)<1$ means that policy samples lie closer to $D_i$ than held-out expert actions do. A decrease in $R_i$ indicates increasing relative proximity to the policy's training split. As shown in Fig.~\ref{fig:split_divergence}(b), $R_i(\tau)$ decreases later in training for both policies. Together with their increasing prediction divergence on $D_3$ in Fig.~\ref{fig:split_divergence}(a), this supports increasingly split-specific fitting.

\paragraph{Observation conditioning exacerbates late-training overfitting.}

To examine the role of observation conditioning, we compare the image-conditioned denoiser $\epsilon_{\theta_\tau}(a_t,t,o)$ with an action-only denoiser $\epsilon_{\gamma_\tau}(a_t,t)$ trained on the same demonstration action chunks. Here $\theta_\tau$ and $\gamma_\tau$ denote the parameters at training epoch $\tau$. The action-only model uses the same denoising backbone and training settings but receives no observation input. We evaluate both models on identical noisy actions from held-out demonstrations. Let $L^m_\tau(t)$ denote the held-out $\epsilon$-prediction MSE at noise level $t$ for $m\in\{\mathrm{cond},\mathrm{uncond}\}$. We measure the relative change from epoch 200 to 800 as
\begin{equation}
    \Delta_{\mathrm{rel}}^m(t)
    = \frac{
        L^m_{{\color{finalblue}800}}(t)
        - L^m_{{\color{selectedorange}200}}(t)
    }{
        L^m_{{\color{selectedorange}200}}(t)
    },
    \qquad m\in\{\mathrm{cond},\mathrm{uncond}\}.
    \label{eq:cond_uncond_heldout_change}
\end{equation}

The ratio is unchanged if $\epsilon$-prediction MSE is converted to clean-action MSE at the same noise level.

\begin{figure}[t]
    \centering
    \includegraphics[width=0.5\linewidth]{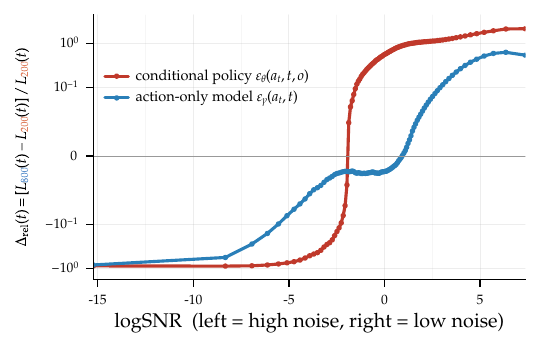}
    \vspace{-0.5em}
    \caption{\textbf{Observation conditioning exacerbates the deterioration in held-out denoising during late training.}
    Relative change in held-out $\epsilon$-prediction MSE from epoch 200 to 800 for the image-conditioned and action-only models, evaluated on the same noisy action chunks. Negative values indicate lower held-out loss at epoch 800; positive values indicate higher loss. The conditional model exhibits held-out loss deterioration at higher diffusion noise levels.}
    \label{fig:cond_uncond_deterioration}
\end{figure}

Fig.~\ref{fig:cond_uncond_deterioration} shows that $\Delta_{\mathrm{rel}}^m(t)$ changes from negative to positive as noise decreases for both models. Continued training therefore improves held-out denoising at high noise but worsens it near the clean end. The reversal occurs at higher noise for the conditional model, which also shows substantially greater deterioration toward low noise. One possible explanation is that as noise decreases, the noisy action becomes more informative about individual training actions. Observation provides an additional identifier for training examples, potentially enabling memorization when the action is more heavily corrupted.

We next examine how the conditional model's late-training deterioration depends on the \emph{density} of training observations near a held-out query. We approximate this density using the ground-truth simulator state because the policy is image-conditioned. For a query state $s$ and training episode $e$, define
\begin{equation}
    d(s,e)=\min_{s'\in e}\|s-s'\|_2.
    \label{eq:episode_distance}
\end{equation}
Let $r_{10}(s)$ be the 10th-smallest $d(s,e)$ across the 200 training episodes. We divide held-out queries into dense, medium, and sparse thirds according to $r_{10}$, with larger values indicating sparser support.

For each group $g\in\{\mathrm{dense},\mathrm{medium},\mathrm{sparse}\}$, let $L^{\mathrm{cond}}_{\tau,g}(t)$ be the conditional model's mean held-out $\epsilon$-prediction MSE over queries in that group. Using the same relative-change measure as Eq.~\ref{eq:cond_uncond_heldout_change}, we compute
\begin{equation}
    \Delta_{\mathrm{rel},g}(t)
    = \frac{
        L^{\mathrm{cond}}_{{\color{finalblue}800},g}(t)
        - L^{\mathrm{cond}}_{{\color{selectedorange}200},g}(t)
    }{
        L^{\mathrm{cond}}_{{\color{selectedorange}200},g}(t)
    }.
    \label{eq:delta_loss}
\end{equation}

\begin{figure*}[t]
    \centering
    \includegraphics[width=0.5\linewidth]{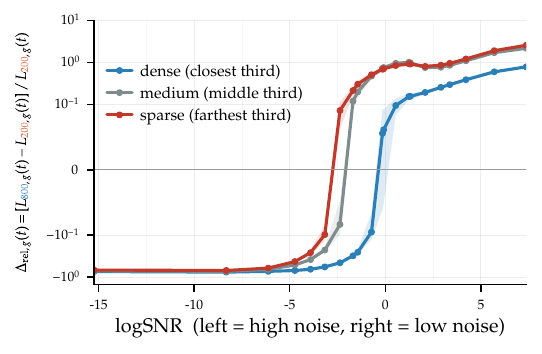}
    \vspace{-0.5em}
    \caption{\textbf{Late-training held-out denoising loss increases more at lower diffusion noise levels and with sparser local conditioning support.}
    Relative change in mean held-out $\epsilon$-prediction MSE from epoch ${\color{selectedorange}200}$ to epoch ${\color{finalblue}800}$, $\Delta_{\mathrm{rel},g}(t)$ (Eq.~\ref{eq:delta_loss}), for queries divided into equal thirds by local conditioning support. Negative values indicate lower mean loss at epoch 800; positive values indicate higher mean loss. At a fixed noise level, the relative loss change is generally larger for queries with sparser local conditioning support.}
    \label{fig:condition_support_overfit}
\end{figure*}

Fig.~\ref{fig:condition_support_overfit} shows that the reversal from improvement to deterioration starts from higher noise levels for sparsely supported observations than for densely supported ones. Near the clean end, the relative increase in held-out loss is also generally larger in the less-supported groups. Thus, the noise range in which continued training harms generalization depends on how well the observation is covered by the training demonstrations.

\paragraph{Local-support analysis.}
To relate this pattern to finite-data support, we count training examples compatible with a denoising query in both observation conditioning and action space.

Assume the training windows
$\mathcal D_n=\{(Z_i,A_i)\}_{i=1}^n$ are i.i.d. samples from a joint density $p_{Z,A}$ that is continuous near the held-out query. Here $Z_i\in\R^{d_z}$ is a conditioning representation and $A_i\in\R^{d_a}$ is the clean expert action. Take an independent held-out pair $(Z^\star,A^\star)=(z^\star,a^\star)$ and one initial diffusion noise draw $\eps$.
At diffusion time $t$,
\begin{equation}
    x_t=\alpha_t a^\star+\sigma_t\eps,
    \qquad
    r_t:=\frac{\sigma_t}{\alpha_t},
    \qquad
    \widetilde a_t:=\frac{x_t}{\alpha_t}=a^\star+r_t\eps.
    \label{eq:query_defs_simple}
\end{equation}

For a candidate clean training action $A_i$, the forward diffusion process gives
\begin{align}
    p(x_t\mid A_i)
    &\propto
    \exp\!\left(-\frac{\|x_t-\alpha_t A_i\|^2}{2\sigma_t^2}\right) \\
    &=
    \exp\!\left(-\frac{\|A_i-\widetilde a_t\|^2}{2r_t^2}\right).
    \label{eq:action_likelihood_simple}
\end{align}

Eq.~\ref{eq:action_likelihood_simple} shows that the likelihood depends on $\|A_i- \widetilde a_t\|_2/r_t$. An action at distance $c r_t$ from $\widetilde a_t$ has relative likelihood $e^{-c^2/2}$. Therefore, for any fixed relative-likelihood threshold, the radius of the corresponding set of clean actions is proportional to $r_t$. As $r_t$ decreases, this set shrinks around $\widetilde a_t$.

We now introduce a simple local-support model. Fix a radius $h$ in conditioning space, and replace the smooth Gaussian weighting in Eq.~\ref{eq:action_likelihood_simple} with a hard action-space cutoff at $r_t$. We then count training examples that satisfy both conditions:
\begin{equation}
N_{\mathrm{loc}}(t)
=
\sum_{i=1}^n
\mathbf 1\!\{\|Z_i-z^\star\|_2\le h\}
\mathbf 1\!\{\|A_i-\widetilde a_t\|_2\le r_t\}.
\label{eq:nloc_def}
\end{equation}
Thus, an example is counted if its conditioning representation lies within
distance $h$ of $z^\star$ and its clean action lies within distance $r_t$ of
$\widetilde a_t$. The cutoffs define an analytical neighborhood, not which examples the network actually uses. 

Let $q_t$ be the probability that one i.i.d. training pair is counted by
Eq.~\ref{eq:nloc_def}. Then
\begin{align}
q_t
&=\Pr\!\left(
\|Z-z^\star\|\le h,
\|A-\widetilde a_t\|\le r_t
\right) \\
&=\int_{\|z-z^\star\|\le h}
   \int_{\|a-\widetilde a_t\|\le r_t}
   p_{Z,A}(z,a)\,da\,dz.
\label{eq:qt_start}
\end{align}
It follows that
$N_{\mathrm{loc}}(t)\sim\mathrm{Binomial}(n,q_t)$ and
$\E[N_{\mathrm{loc}}(t)]=nq_t$.

To evaluate $q_t$, substitute $z=z^\star+h u$ and
$a=\widetilde a_t+r_t v$:
\begin{equation}
q_t
=h^{d_z}r_t^{d_a}
\int_{\|u\|\le1}\int_{\|v\|\le1}
 p_{Z,A}(z^\star+h u,\widetilde a_t+r_t v)\,dv\,du.
\label{eq:qt_change}
\end{equation}
Near the low-noise end, \(r_t\to 0\) and \(\widetilde a_t=a^\star+r_t\eps\to a^\star\). In the joint limit \(h,r_t\to 0\), assuming \(p_{Z,A}\) is continuous and positive at \((z^\star,a^\star)\), Eq. \(\ref{eq:qt_change}\) gives
\begin{equation}
\E[N_{\mathrm{loc}}(t)]
=
V_{d_z}V_{d_a}\,n p_{Z,A}(z^\star,a^\star)
 h^{d_z}r_t^{d_a}[1+o(1)],
\label{eq:nloc_expectation}
\end{equation}
where $V_d$ is the volume of the unit ball in $\R^d$.

Factoring
$p_{Z,A}(z^\star,a^\star)=p_Z(z^\star)
 p_{A\mid Z}(a^\star\mid z^\star)$ gives
\begin{equation}
\E[N_{\mathrm{loc}}(t)]
\approx Cn
\underbrace{p_Z(z^\star)h^{d_z}}
_{\text{local conditioning mass}}
\underbrace{p_{A\mid Z}(a^\star\mid z^\star)r_t^{d_a}}
_{\text{local action mass at noise scale }r_t},
\label{eq:nloc_factorization}
\end{equation}
with $C=V_{d_z}V_{d_a}$. Sparser local conditioning support lowers the first factor; lower diffusion noise reduces $r_t=\sigma_t/\alpha_t$ and hence the second factor. Both lead to fewer locally compatible training examples. This reduced support is consistent with greater vulnerability to finite-data overfitting, but the calculation itself does not predict when a trained denoising network's held-out loss will increase.

The $r_{10}$ measure used in Fig.~\ref{fig:condition_support_overfit} is computed in simulator-state space, rather than the policy's learned conditioning representation, and serves as a proxy for local conditioning support.

\paragraph{Low-noise changes affect closed-loop pass@$k$.}
To test whether changes to low-noise denoising affect closed-loop success, we divide the 10 DDIM inference calls into five consecutive two-call bands, Q1--Q5, and replace the late checkpoint's weights with the early checkpoint's weights for one band at a time. Patching Q1--Q4 has little effect on pass@$1$ or pass@$10$, whereas replacing Q5 recovers most of the earlier checkpoint's pass@$10$ advantage while reducing pass@$1$ (Fig.~\ref{fig:app_noise_resolved_patching}(c)). Thus, for this checkpoint pair, the final two calls account for most of the observed differences in the pass@$k$ profile.
\clearpage
\section{Full Coverage--Reliability Results and Checkpoint Selection}
\label{app:coverage_reliability_full_results}

We report the complete results across five MimicGen tasks with 200 and 500
demonstrations, together with multitask RoboCasa pretraining. For each setting,
we show best-of-$k$ validation loss, closed-loop pass@$k$, the full error-quantile
profile, coverage--reliability checkpoint selection, and downstream residual RL or supervised finetuning.
\vspace{0.5em}

\begin{center}
    \includegraphics[width=\textwidth]
    {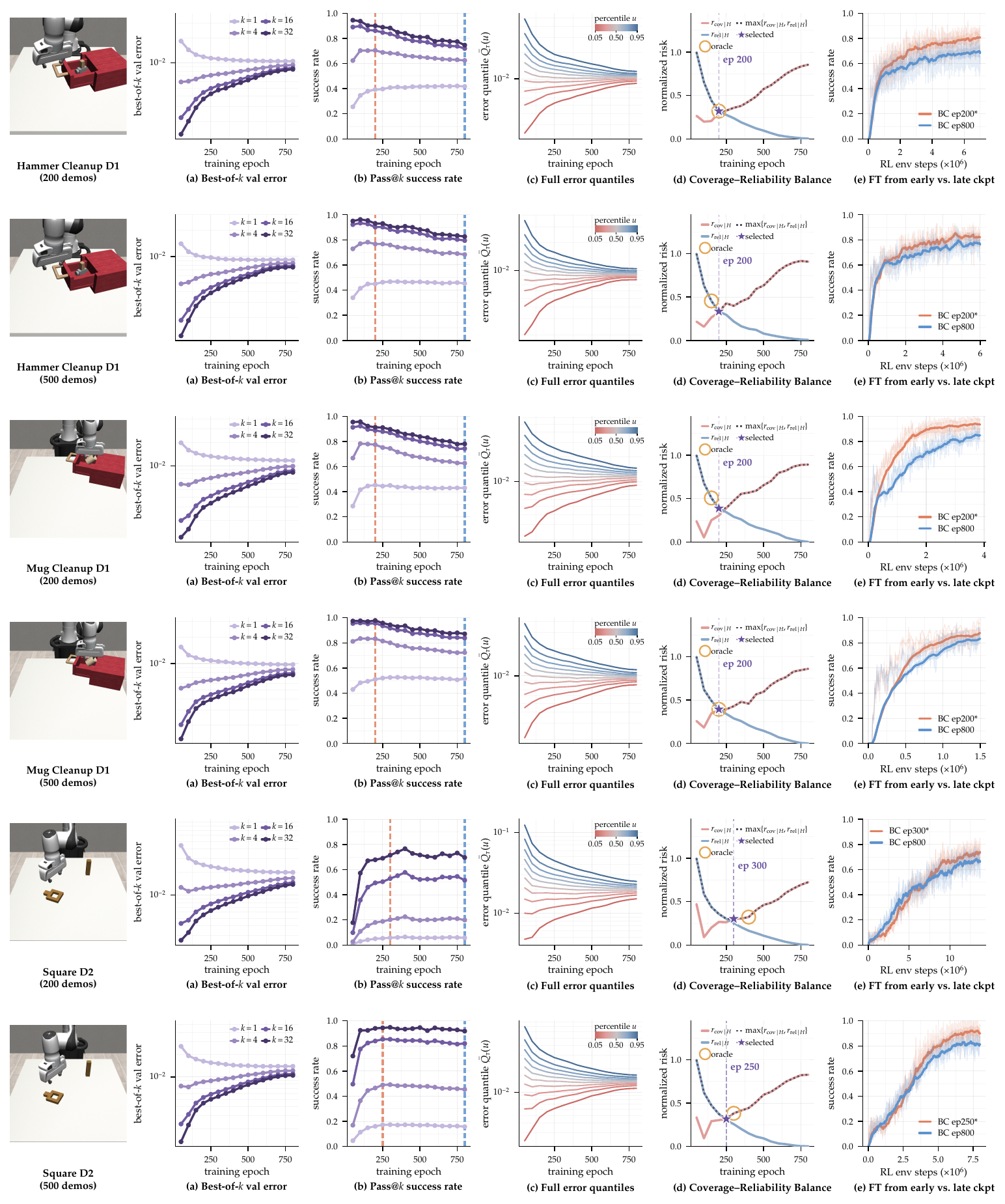}
\end{center}

\clearpage

\begin{center}
    \includegraphics[width=\textwidth]
    {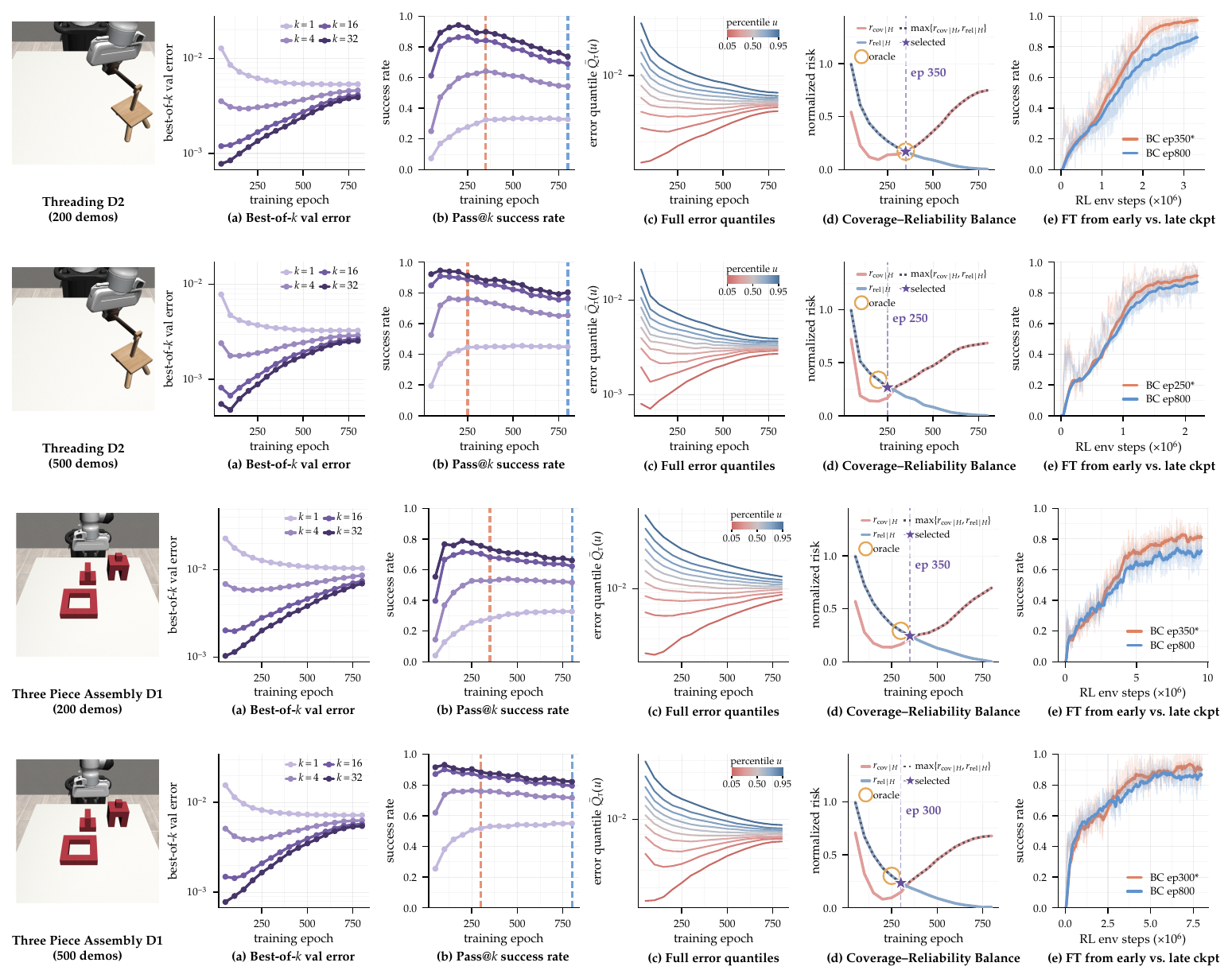}

    \captionof{figure}{
    \textbf{Full MimicGen results across tasks and dataset sizes.}
    Results for all five tasks with 200 and 500 demonstrations.
    Each row reports best-of-$k$ validation loss, closed-loop pass@$k$,
    the full error-quantile profile, coverage--reliability checkpoint
    selection, and downstream residual RL finetuning.
    }
    \label{fig:app_mimicgen_full}
\end{center}

\begin{center}
    \includegraphics[width=\textwidth]
    {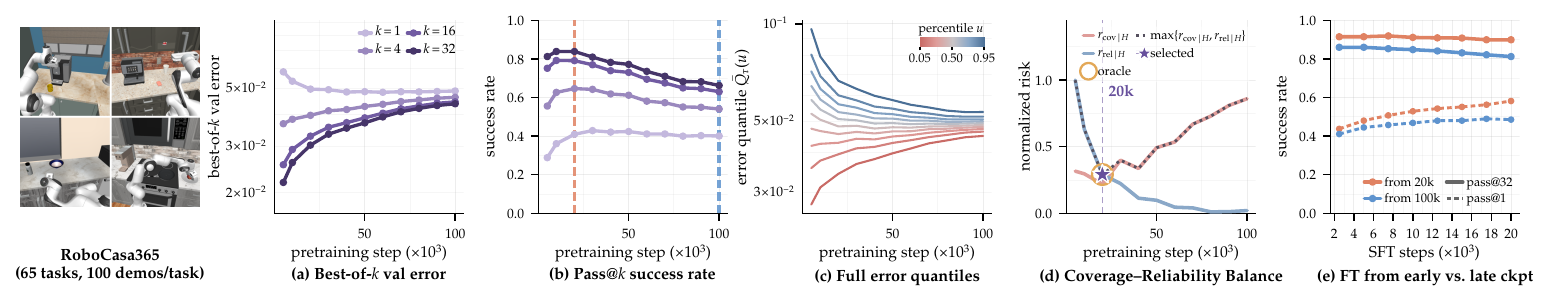}

    \captionof{figure}{
    \textbf{Full RoboCasa multitask pretraining results.}
    Best-of-$k$ validation loss, pass@$k$, error quantiles,
    coverage--reliability checkpoint selection, and downstream SFT.
    }
    \label{fig:app_robocasa_full}
\end{center}

\paragraph{The same learning dynamics hold across tasks and dataset sizes.}
Figures~\ref{fig:app_mimicgen_full} and~\ref{fig:app_robocasa_full} show a consistent pattern across all settings. Best-of-$k$ loss and pass@$k$ favor increasingly earlier checkpoints as $k$ increases, while the error-quantile curves exhibit the same tradeoff between favorable-tail coverage and average-case reliability as described in Sec.~\ref{sec:offline_diagnosis}. The pattern holds for both single-task MimicGen BC and multitask RoboCasa pretraining. Importantly, pass@$1$ and $k{=}1$ best-of-$k$ loss (mean held-out action MSE) generally favor later checkpoints after large-$k$ pass@$k$ has begun to decline.

Increasing the MimicGen dataset from 200 to 500 demonstrations generally attenuates, and in several cases delays, the late-stage loss of coverage. While two discrete dataset sizes are insufficient for a scaling-law claim, this trend is consistent with the finite-data overfitting diagnosis in Appendix.~\ref{app:overfitting_diagnosis}.

\paragraph{Coverage--reliability balance criterion agrees with a closed-loop oracle.}
We apply the coverage--reliability criterion from Sec.~\ref{sec:offline_diagnosis} to every setting. For comparison, we define the closed-loop oracle directly from the full pass@$k$ table. Here we describe how we compute the checkpoint selected from oracle closed-loop pass@$k$ runs. For each $k$, we normalize pass@$k$ across checkpoints as
\begin{equation}
g_\tau(k)
=
\frac{
\max_{\tau'} P_k(\tau') - P_k(\tau)
}{
\max_{\tau'} P_k(\tau') - \min_{\tau'} P_k(\tau')
},
\label{eq:closed_loop_regret}
\end{equation}
where $P_k(\tau)$ denotes pass@$k$ at checkpoint $\tau$. We then select
\begin{equation}
\tau_{\mathrm{oracle}}
=
\arg\min_\tau \max_k g_\tau(k).
\label{eq:closed_loop_oracle}
\end{equation}
This oracle selects the checkpoint whose worst normalized gap from the best checkpoint is smallest across the full pass@$k$ profile. Across tasks and dataset sizes, the checkpoint selected by our proposed offline criterion lies near the closed-loop oracle.

\paragraph{Selected checkpoints also support stronger finetuning.}
For all MimicGen settings, residual RL initialized from the selected checkpoint outperforms RL from the final BC checkpoint. RoboCasa shows the same pattern under SFT: initialization from the selected pretraining checkpoint yields stronger pass@1 and pass@32 throughout finetuning. These results extend the main-text finding that the coverage--reliability balance retained during supervised training leads to stronger finetuning performance compared to the final checkpoint from the same run.

\begin{figure*}[!t]
    \centering
    \includegraphics[width=\textwidth]
    {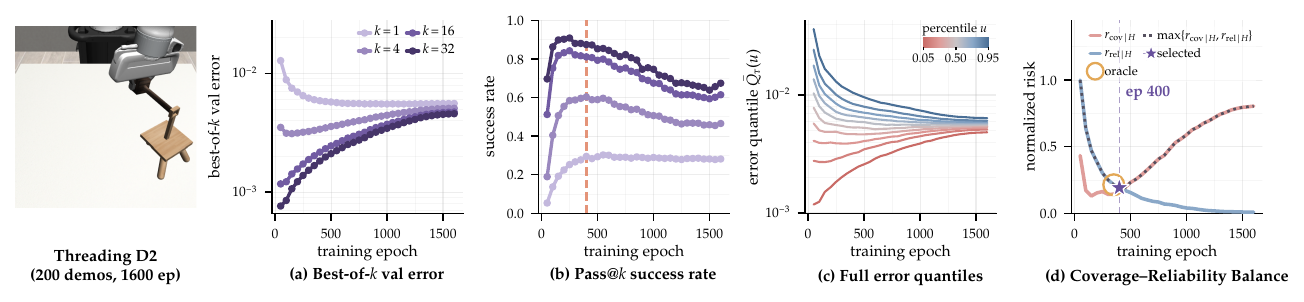}
    \caption{\textbf{Extended training on Threading D2.}
    The coverage--reliability dynamics persist to 1600 epochs, and the offline
    criterion continues to select an intermediate checkpoint.}
    \label{fig:app_threading_1600}
\end{figure*}

\paragraph{Selection is robust to training horizon.}
We additionally extend the 200-demonstration Threading D2 run from 800 to 1600 epochs. Figure~\ref{fig:app_threading_1600} shows that the same pass@$k$ and error-quantile dynamics persist under substantially longer training, while the criterion continues to select an intermediate checkpoint.

\begin{wrapfigure}{r}{0.34\columnwidth}
    \centering
    \vspace{-0.8em}
    \includegraphics[width=\linewidth]
    {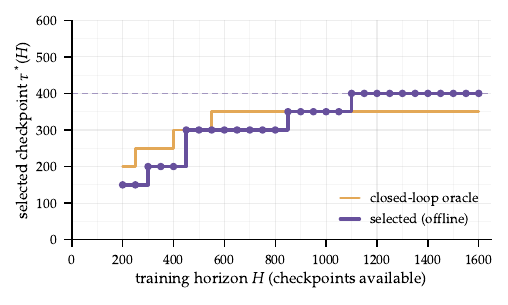}
    \caption{\textbf{Robustness to training horizon.}
    Offline selection tracks the closed-loop oracle as $H$ increases.}
    \label{fig:app_selection_horizon}
    \vspace{-0.8em}
\end{wrapfigure}

We further recompute both selectors using only checkpoints available up to each training horizon $H$. Figure~\ref{fig:app_selection_horizon} shows that the offline-selected checkpoint closely tracks the corresponding closed-loop oracle as additional checkpoints become available, rather than simply moving toward the end of training. The selection criterion therefore does not depend sensitively on the chosen training horizon.
\section{Additional Results on \JointOA}
\label{app:jointaug_supp}

\appsubsec{Additional Task}
\label{app:jointaug_additional_task}

\begin{figure}[t]
    \centering
    \includegraphics[width=\linewidth]
    {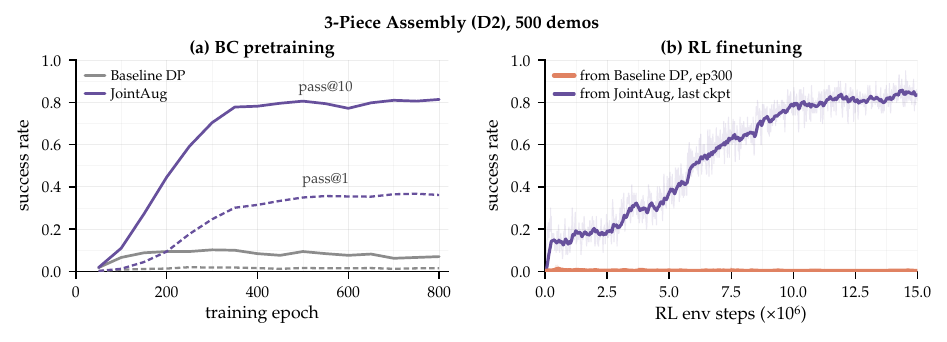}
    \caption{\textbf{\JointOA{} enables residual RL finetuning on 3-Piece Assembly (D2).}
    \emph{Left:} \JointOA{} substantially improves pass@$1$ and pass@$10$ during BC pretraining.
    \emph{Right:} residual RL from the final \JointOA{} checkpoint reaches over $0.8$ success, while finetuning from the baseline checkpoint remains near zero.}
    \label{fig:app_jointaug_three_piece}
\end{figure}

We additionally evaluate \JointOA{} on 3-Piece Assembly (D2) with 500 demonstrations. This task provides a more extreme example of the effect of pretraining coverage on residual RL finetuning. As shown in Fig.~\ref{fig:app_jointaug_three_piece}, \JointOA{} raises pass@$10$ to roughly $0.8$ during BC pretraining, compared with below $0.1$ for
the baseline. Moreover, residual RL from the final \JointOA{} checkpoint then reaches over $0.8$ success, while finetuning from the baseline checkpoint remains near zero throughout the evaluated interaction budget. Thus, on this task, the pretraining difference changes residual RL from not working to successful finetuning.

\appsubsec{Robustness to Generator Checkpoint Selection}
\label{app:jointaug_robust_to_ckpt}

\begin{figure*}[t]
    \centering
    \includegraphics[width=\textwidth]
    {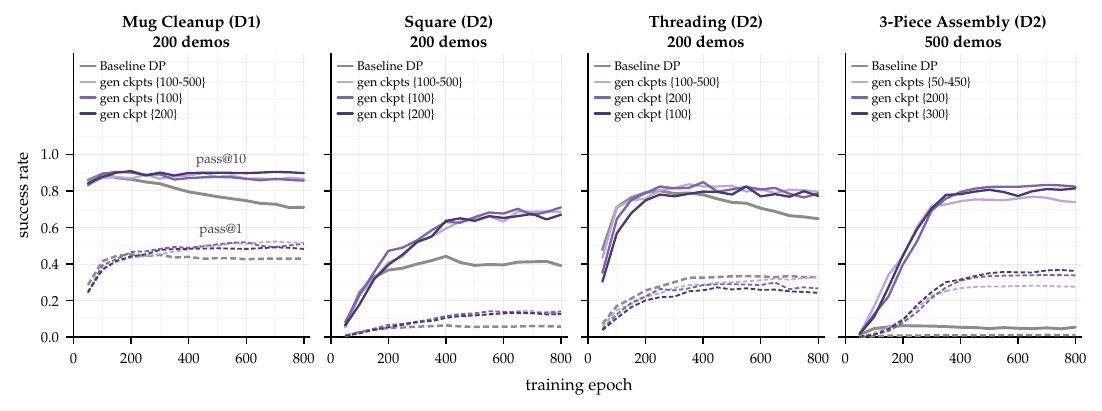}
    \caption{\textbf{\JointOA{} is relatively robust to generator checkpoint selection.}
    We compare the five-checkpoint mixture used in our main experiments against individual generator checkpoints from the same run . Across four tasks, single-checkpoint variants have similar pass@$10$ gains over the baseline, while the mixture performs similarly overall and slightly improves pass@$1$ on \texttt{Mug Cleanup} and \texttt{Threading}.}
    \label{fig:app_jointaug_ckpt_robustness}
\end{figure*}

By default, \JointOA{} uses a a uniform mixture of five generator checkpoints, spaced 100 epochs apart, around the transition identified by our train--held-out nearest-neighbor ratio (Appendix~\ref{app:impl_jointaug}). We use this short checkpoint window to average supervision across the generator's transition from population-level generalization toward increasing training-data bias.

To test how sensitive \JointOA{} is to this heuristic choice, we compare the five-checkpoint mixture against distilling from individual generator checkpoints from the same region. Fig.~\ref{fig:app_jointaug_ckpt_robustness} shows that the resulting policy curves are similar across these choices on all four tasks. In particular, single generator checkpoints preserve most of the pass@$10$ improvement over the baseline. The five-checkpoint mixture performs similarly overall and gives slightly higher pass@$1$ on \texttt{Mug Cleanup} and \texttt{Threading}. Together, these results show that \JointOA{} does not depend strongly on selecting one particular generator checkpoint or checkpoint mixture. Mixing several nearby checkpoints provides a simple way to reduce checkpoint sensitivity.

\appsubsec{Visualizing How \JointOA{} Changes the Action Distribution}
\label{app:jointaug_action_cloud_comp}

\begin{figure*}[t]
    \centering
    \includegraphics[width=\textwidth]
    {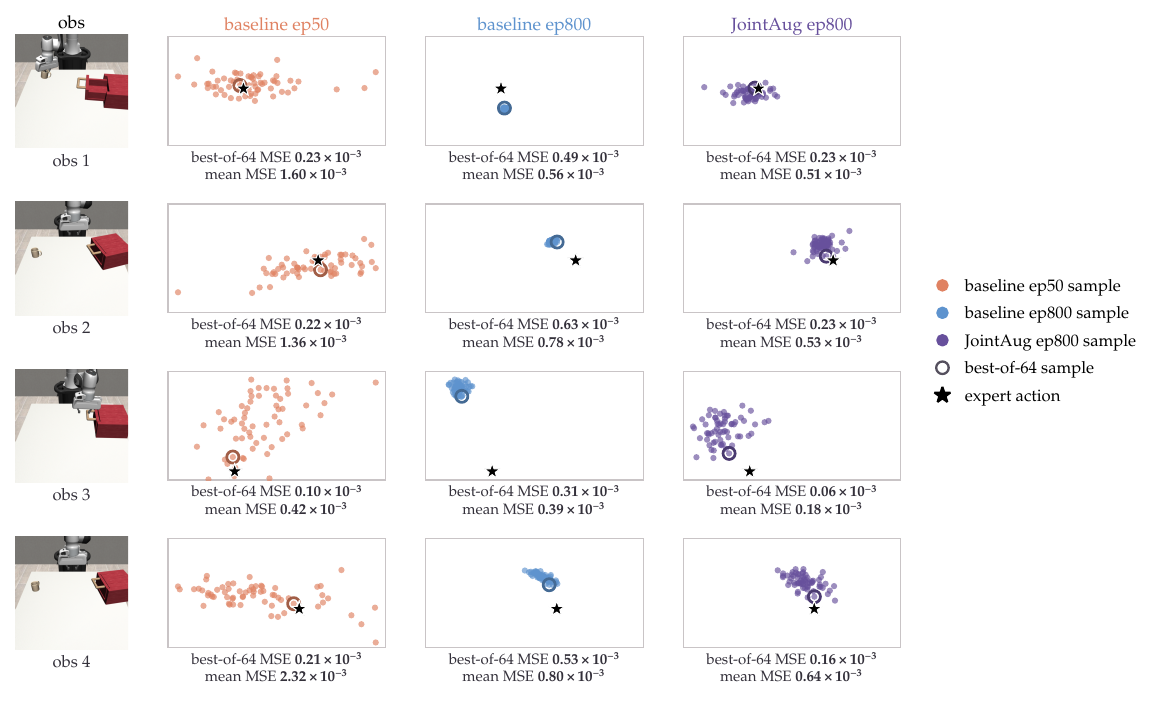}
    \caption{\textbf{\JointOA{} retains favorable-tail coverage while improving
    typical action accuracy.}
    For four held-out \texttt{Mug Cleanup} states, we jointly project 64 action samples from \textcolor{selectedorange}{baseline ep50}, \textcolor{finalblue}{baseline ep800}, and \JointOA{} ep800 with PCA.
    Stars mark expert actions and rings mark the minimum-MSE sample.
    The \textcolor{selectedorange}{baseline ep50} is dispersed with low best-of-64 but high mean MSE;
    baseline \textcolor{finalblue}{baseline ep800} contracts and lowers mean MSE but worsens best-of-64 MSE;
    \JointOA{} ep800 achieves best of both worlds.}
    \label{fig:appendix_action_cloud_mug}
\end{figure*}

To understand why \JointOA{} improves both coverage and reliability, we inspect the action distribution at held-out observations. For each state, we sample 64 action chunks from an early baseline checkpoint (ep50), the final baseline checkpoint (ep800), and the final \JointOA{} checkpoint (ep800), and jointly project the samples with PCA. We report both the mean action MSE and the best-of-64 MSE to the expert action. The former measures the accuracy of a typical sample, while the latter measures whether the sampled distribution still contains an action close to the expert.

Fig.~\ref{fig:appendix_action_cloud_mug} shows that the early baseline checkpoint produces a broad action distribution: across the four states, best-of-64 MSE is very small, but mean MSE can go up to $2.32\times10^{-3}$. Thus, lucky draw samples lie very close to the expert, while the majority of samples are much farther away. This provides coverage but poor reliability for stable closed-loop execution.

By ep800, the baseline distribution contracts substantially. Mean MSE improves a lot from the early checkpoint, but best-of-64 MSE worsens on all four states. The later baseline therefore produces more consistent samples, but its best samples are farther from the expert, thus losing coverage. 

The final \JointOA{} policy checkpoint combines the favorable properties of both checkpoints: its action clouds remain broader than the final baseline while being centered closer to the expert. Mean MSE falls further to $0.18$--$0.64\times10^{-3}$, and best-of-64 MSE decreases to $0.06$--$0.23\times10^{-3}$. Thus, \JointOA{} achieves lower mean and best-of-64 error than the final baseline on every state, while nearly matching or improving the favorable tail of the early baseline.

Together, these visualizations provide a post-hoc explanation for the pass@1 and large $k$ pass@$k$ improvement from \JointOA{}. Jointly generated $(o,a)$ pairs expand supervision beyond the finite demonstrations while preserving observation--action correspondence. The resulting policy retains a broader set of low-error actions without the large average errors of the early baseline.

\appsubsec{Additional Baseline - Context-Smoothed Pre-training in \TMRL{}}
\label{app:additional_baseline_tmrl}

\begin{figure*}[t]
    \centering
    \includegraphics[width=\textwidth]
    {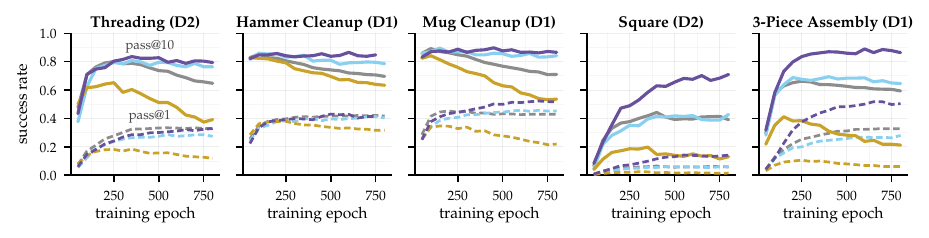}
    \vspace{-0.4cm}

    \includegraphics[width=0.56\textwidth]
    {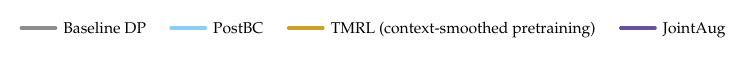}
    \vspace{-0.4cm}

    \caption{\textbf{Context-Smoothed Pre-training from \TMRL{} does not preserve coverage under prolonged training.}
    Across five MimicGen tasks, \TMRL{} generally underperforms \JointOA{} in pass@$10$ and often loses both pass@$10$ and pass@$1$ at later checkpoints.}
    \label{fig:app_tmrl_baseline}
\end{figure*}

We additionally compare against the context-smoothed pretraining part from \TMRL{}~\citep{hong2026tmrl}, which is designed to increase action coverage by reducing the amount of information available in the policy context. During pretraining, \TMRL{} applies a forward diffusion noise to the observation
context,
\[
q(\tilde c_t \mid c)
=
\mathcal{N}
\left(
\sqrt{\bar{\alpha}_t}\,c,\,
(1-\bar{\alpha}_t)I
\right),
\]
then conditions the policy on both the corrupted context $\tilde c_t$ and its corresponding noise level $t$. The intuition is that increasing $t$ aliases a wider neighborhood of contexts, so the corresponding action distribution can borrow actions from other demonstrations rather than fitting only the conditional distribution at the original observation. We tune the maximum context-noise level for each task and report the best-performing setting; full implementation details are in Appendix~\ref{app:impl_tmrl}.

Fig.~\ref{fig:app_tmrl_baseline} shows that context smoothing does not preserve coverage well in our setting. Across tasks, its pass@$10$ remains below \JointOA{} and often declines substantially with further training; its pass@$1$ also tends to degrade

One likely reason is that context smoothing relies on local structure in the latent space where noise is applied. Isotropic Gaussian perturbations are useful when nearby context representations correspond to observations with compatible actions, since smoothing can then borrow coherent behaviors from neighboring
contexts. In our image-based diffusion policies, however, this assumption may not hold. The observation encoder is trained jointly from task demonstrations and is not explicitly regularized so that Euclidean neighborhoods in its latent space preserve action similarity. Context corruption can therefore mix observations that require different actions. We note that differs from the image-based experiments in \citet{hong2026tmrl}, which smooth pretrained VLM embeddings.

More generally, context smoothing increases overlap between conditional action distributions, but by design does not directly constrain the favorable tail of the action distribution as supervised training proceeds. It therefore does not guarantee preservation of large-$k$ pass@$k$.
\section{Additional Results and Visualizations}
\label{app:additional_results}
\appsubsec{Sampling Configuration Ablations}
\label{app:sampling_config_ablations}
\begin{figure}[t]
    \centering
    \includegraphics[width=0.5\columnwidth]
    {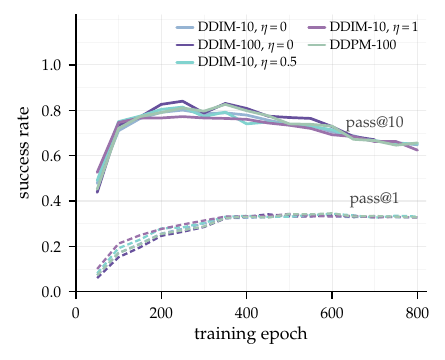}
    \caption{\textbf{Pass@$k$ training dynamics are stable across sampling
    configurations.}
    Pass@$1$ and pass@$10$ are evaluated using DDIM with different numbers of
    denoising steps and $\eta$, and using DDPM. All settings show the same
    late-training decline in pass@$10$.}
    \label{fig:app_sampling_ablation}
\end{figure}

Our main experiments use DDIM sampling with 10 denoising steps and $\eta=0$. The coverage--reliability tradeoff develops during diffusion-policy training rather than being induced by a particular sampling procedure. Fig.~\ref{fig:app_sampling_ablation} compares DDIM with 10 or 100 denoising steps, $\eta\in\{0,0.5,1\}$, and DDPM with 100 steps. Across all settings, pass@$10$ peaks earlier and declines with further training, while pass@$1$ continues to improve or remains stable. The same trend across sampling configurations shows that the late-training coverage loss is a property of the trained denoising model rather than the sampling configuration used for evaluation.

\appsubsec{Learning-rate Schedule Ablations}
\label{app:lr_schedule_ablations}

Our main experiments show that large-$k$ pass@$k$ peaks earlier in supervised pretraining than pass@$1$, and can decline as pass@$1$ continues to improve. Because the learning-rate schedule controls how aggressively the policy is updated over training, it can affect when these changes occur. Related work on diffusion models likewise finds that optimization settings can shift the onset of finite-data fitting~\citep{bonnaire2026diffusion,wu2025taking}. We therefore test whether changing the learning-rate schedule fundamentally changes the coverage--reliability tradeoff itself, or primarily shifts the timescales at which it appears. Across the schedules we evaluate, the same qualitative pattern remains: pass@$10$ peaks earlier and declines at later epochs, while pass@$1$ improves over a longer portion of training.

\textbf{Setup.}
For single-task MimicGen, our main experiments follow the optimizer configuration of the public Diffusion Policy implementation~\citep{chi2025diffusion}: AdamW with learning rate $10^{-4}$, $\beta_1=0.95$, $\beta_2=0.999$, weight decay $10^{-6}$, and 500 warmup steps followed by a single half-cosine decay from $10^{-4}$ to zero over the full training horizon. We use this standard recipe throughout experiments included in the main text. 

We repeat the 200-demonstration \texttt{Threading 2D} experiment with four alternative schedules while keeping all other training settings fixed. (1) Constant $10^{-4}$, which removes annealing while preserving the maximum learning rate of the default schedule; (2) constant $5\times10^{-5}$ to test whether reducing the learning rate shifts both the improvement in pass@$1$ and the decline in large $k$ pass@$k$; (3) cosine decay from $10^{-4}$ to $10^{-5}$ which preserves the monotone decay of the default schedule but removes its near-zero learning rates late in training; (4) periodic cosine with four cycles to test whether repeatedly increasing and decreasing the learning rate changes the same trend or instead produces different behavior.

\begin{figure*}[t]
    \centering
    \includegraphics[width=\textwidth]{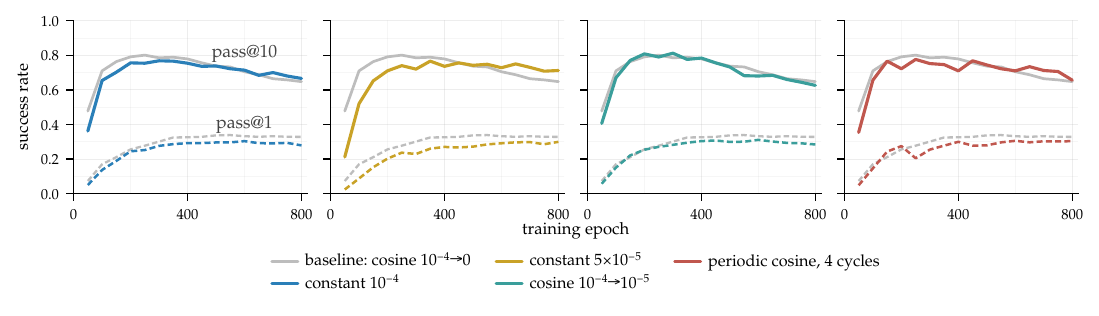}
    \caption{\textbf{Learning-rate schedule ablation on \texttt{Threading D2}}
    Each panel overlays one ablated schedule (colored) with the default single-cosine schedule (gray).
    Solid and dashed curves denote pass@$10$ and pass@$1$, respectively.
    Across schedules, large-$k$ pass@$k$ peaks earlier and declines as pass@$1$ improves later in training; changing the learning-rate schedule primarily shifts the timing of these changes.}
    \label{fig:app_lr_schedule}
\end{figure*}

Together, these settings cover fixed, monotone-decay, and cyclic learning-rate schedules, while isolating the effects of learning-rate magnitude, annealing, and the final learning-rate floor. In Fig.~\ref{fig:app_lr_schedule}, each panel overlays one ablated schedule with the default single cosine schedule (gray); solid and dashed curves denote pass@$10$ and pass@$1$ respectively.

\textbf{Results.}
Removing learning-rate decay doesn't change the decline in large $k$ pass@$k$. With constant $10^{-4}$, pass@$10$ increases early and then decreases over later epochs, while pass@$1$ remains slightly below the default run through most of training. Keeping the learning rate fixed at its maximum value therefore does not preserve the higher late-training pass@$10$ without also changing pass@$1$. Likewise, changing the endpoint of the cosine schedule from zero to $10^{-5}$ produces nearly the same pass@$1$ and pass@$10$ curves as the default schedule. These two comparisons show that the decline in large-$k$ pass@$k$ is not specific to either learning-rate decay or the near-zero learning rates reached at the end of the default schedule.

With constant $5\times10^{-5}$, both metrics change more slowly. Pass@$10$ rises more gradually, reaches its maximum later, and begins declining later than under the default schedule. Pass@$1$ also improves more gradually and remains lower at epoch 800. Thus, the higher pass@$10$ retained at later epochs is accompanied by lower pass@$1$; reducing the learning rate shifts both sides of the tradeoff later in training rather than preserving coverage at the same reliability.

Periodic cosine produces larger cross checkpoint fluctuations but does not change the longer-term trend. Pass@$10$ still reaches its highest values earlier and decreases later in training, while pass@$1$ remains higher at later epochs. Repeated learning-rate cycling therefore makes training less stable across checkpoints without removing the coverage--reliability tradeoff. Among the schedules tested here, the default single-cosine schedule also gives the smoothest learning curves and the highest late-training pass@$1$, supporting its use as the default schedule in our main experiments.

Across all schedules tested, large-$k$ pass@$k$ still peaks earlier and
declines as pass@$1$ improves later in training. Smaller constant learning rates shift both changes to later epochs, while periodic cosine produces larger fluctuations across checkpoints. Thus, the learning-rate schedule changes when the coverage--reliability tradeoff appears, but does not fundamentally escapes it. This mirrors~\citet{bonnaire2026diffusion}, who find that changing the optimizer shifts the onset of generalization and memorization in diffusion models while preserving their ordering. More broadly, \citet{wu2025taking} shows that learning rate can directly regularize finite-data fitting of the diffusion score at small noise.  Designing new optimization strategies that preserve large-$k$ coverage without hurting improvements in pass@$1$ is an exciting direction for future work.

\appsubsec{Denoising Network Size Ablations}
\label{app:network_size_ablations}

\begin{figure}[t]
    \centering
    \includegraphics[width=0.5\columnwidth]
    {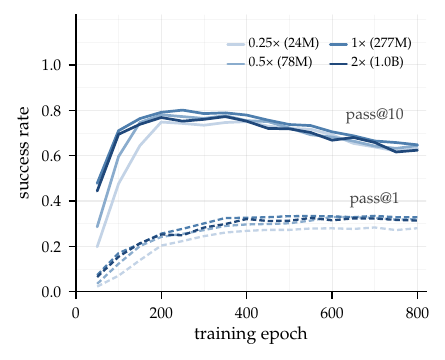}
    \caption{\textbf{The late-training coverage loss persists across denoiser widths.}
    We vary the U-Net from 24M to 1.0B parameters while keeping the encoder, data, optimizer, learning-rate schedule, and sampler fixed.
    All four models show a similar peak and subsequent decline in pass@$10$.}
    \label{fig:app_unet_width}
\end{figure}

We next examine how denoiser network capacity affect the observed learning dynamics between coverage and reliability. We vary the U-Net channel widths while keeping the visual encoder, data, optimizer, learning-rate schedule, and sampler unchanged. We test four denoiser capacities contain 24M, 78M, 277M, and 1.0B parameters. As shown in Fig.~\ref{fig:app_unet_width}, pass@10 peaks between roughly epochs 200--300 for every U-Net model size and then eventually declines to a similar level by epoch 800. Therefore, the loss of coverage during late training is not specific to the capacity of the denoising network. 

\appsubsec{EMA schedule Ablations}
\label{app:ema_schedule_ablations}
EMA changes how quickly the network weights used for evaluation track the current denoiser weights. A possible explanation for the late-training drop in large $k$ pass@$k$ is therefore that the EMA schedule forgets the earlier, higher-coverage weights too quickly. If so, increasing the EMA memory should substantially reduce or remove the decline. Let $\theta_t$ denote the current model parameters and $\bar{\theta}_t$ the EMA parameters. The EMA weights are updated as
\[
\bar{\theta}_t
=
\beta_t \bar{\theta}_{t-1}
+
(1-\beta_t)\theta_t,
\]
where the decay follows the Diffusion Policy warmup schedule
\[
\beta_t
=
\operatorname{clip}\left(
1-\left(1+\frac{s_t}{\gamma^{-1}}\right)^{-p},
\beta_{\min},
\beta_{\max}
\right),
\qquad
s_t=\max(0,t-t_0-1).
\]
Our default uses
$\gamma^{-1}=1$, $p=0.75$, $t_0=0$,
$\beta_{\min}=0$, and $\beta_{\max}=0.9999$.

We ablate over two key hyperparameters, the power $p$ and $\beta_{\max}$, each controling different parts of the EMA memory schedule. Increasing $p$ makes $(1+s_t)^{-p}$ decay faster, so $\beta_t$ approaches one earlier in training. The EMA therefore puts less weight on the current network at each update and retains older weights for longer. In other words, larger $p$ gives the EMA a longer memory earlier in training. By contrast, $\beta_{\max}$ becomes to matter once this growth reaches the cap value. Increasing $\beta_{\max}$ allows $\beta_t$ to move even closer to one before being clipped, giving the EMA a longer maximum memory later in training. 

We compare the default setting $(p=0.75,\beta_{\max}=0.9999)$ against $(0.9,0.9999)$ and $(0.9,0.99995)$. The first comparison increases the EMA memory earlier in training. The second keeps $p$ fixed and increases the maximum memory later in training; for $p=0.9$, the $0.9999$ cap is reached around epoch 309, while the $0.99995$ cap is reached around epoch 668.

\begin{figure}[t]
    \centering
    \includegraphics[width=0.78\columnwidth]
    {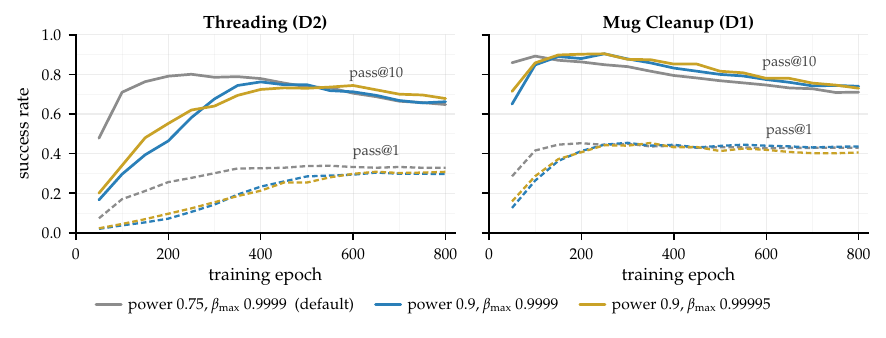}
    \caption{\textbf{Longer EMA memory delays and smooths the pass@$10$ decline but does not fundamentally remove it.}
    Holding $\beta_{\max}=0.9999$ fixed, increasing $p$ from $0.75$ to $0.9$ gives the EMA a longer memory earlier in training, causing pass@$10$ to peak and decline later. Holding $p=0.9$ fixed, increasing $\beta_{\max}$ from $0.9999$ to $0.99995$ allows the EMA memory to continue growing later in training and further smooths the decline. Pass@$10$ still decreases under all schedules.}
    \label{fig:app_ema_schedule}
\end{figure}

Fig.~\ref{fig:app_ema_schedule} shows comparisons between these different temporal averaging schedules. Holding $\beta_{\max}=0.9999$ fixed, increasing $p$ from $0.75$ to $0.9$ gives the EMA a longer memory earlier in training. Accordingly, the $p=0.9$ policy lags more while the underlying current model is improving: pass@$10$ rises more slowly and peaks later. Its decline after the peak is also more gradual because the EMA continues to place more weight on parameters from earlier checkpoints.

Holding $p=0.9$ fixed isolates the effect of $\beta_{\max}$. The two schedules are identical until the $\beta_{\max}=0.9999$ run reaches its cap; after that, $\beta_{\max}=0.99995$ allows the EMA memory to keep growing and therefore retains earlier parameters for longer. This further smooths the late-training change in pass@$10$ for a little bit, but does not fundamentally prevent its decline. Thus, increasing either the early growth or the maximum length of the EMA memory can delay and smooth the loss of coverage, but neither prevents it.

\appsubsec{Isolating Coverage Loss to the Denoising Network}
\label{app:low_dim}
To determine how much of the late-training coverage loss can be attributed to the visual encoder or the denoising network, we run three complementary experiments.
(1) \emph{Freeze the encoder}: we freeze the visual encoder at epoch 150 and continue training only the denoiser, testing whether removing further encoder updates changes the subsequent coverage decline.
(2) \emph{Fresh denoiser with a fixed encoder}: we train a new denoiser from scratch while keeping either the epoch-150 or epoch-800 encoder fixed throughout training. This isolates whether denoiser optimization alone can reproduce the peak-then-decline pattern under a fixed representation.
(3) \emph{Low-dimensional observations}: we train diffusion policies directly from state observations, removing the visual encoder entirely. This tests whether the same coverage--reliability tradeoff arises without any learned visual representations.

\begin{figure}[t]
    \centering
    \includegraphics[width=0.60\columnwidth]
    {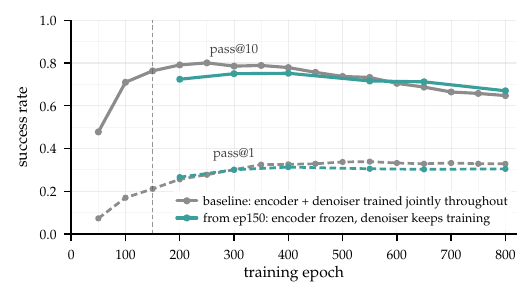}
    \caption{\textbf{Freezing the encoder does not prevent the late-training large $k$ pass@$k$ decline.}
    From epoch 150 onward, the encoder is fixed and only the denoiser is trained. Pass@$10$ still decreases at later epochs. The overall training dynamics is largely the same as baseline run.}
    \label{fig:app_encoder_c1}
\end{figure}

\textbf{Freezing the encoder does not stop the coverage decline.}
We choose epoch 150 according to a split-data representation comparison shows that the nearest-neighbor structure of independently trained encoders has stabilized by this point. Starting from the epoch-150 checkpoint, we replace the encoder with its EMA copy, freeze all encoder parameters, and continue training the denoiser till the end of the training horizon. Fig.~\ref{fig:app_encoder_c1} shows that pass@$10$ still declines at later epochs, with a similar pass@$1$ trend to the original run. Continued encoder updates are therefore not responsible for the coverage loss.

\begin{figure}[t]
    \centering
    \includegraphics[width=0.60\columnwidth]
    {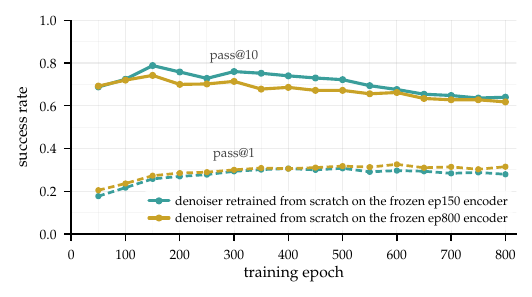}
    \caption{\textbf{Fresh denoisers reproduce the pass@$10$ decline with fixed encoder representations.}
    New denoisers are trained from scratch with either the epoch-150 or epoch-800 encoder frozen throughout training. Both show a later decline in large $k$ pass@$k$.}
    \label{fig:app_encoder_c2}
\end{figure}

\textbf{A fresh denoiser reproduces the decline with a frozen learned visual representation.}
We next randomly initialize a new denoiser and train it from scratch for 800 epochs while keeping the encoder fixed throughout training. We run this experiment with both the epoch-150 and epoch-800 encoders, using the same denoiser initialization. As shown in Fig.~\ref{fig:app_encoder_c2}, pass@$10$ first increases and then declines for both frozen representatioons. The decline therefore reappears even when the encoder never changes and the denoiser is trained entirely from scratch.

\begin{figure*}[t]
    \centering
    \includegraphics[width=\textwidth]
    {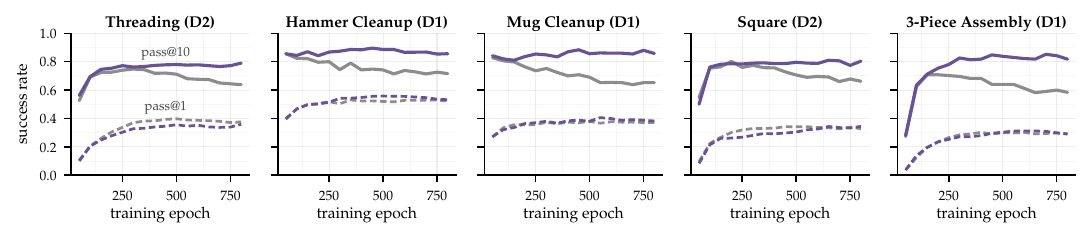}
    \vspace{-0.4cm}

    \includegraphics[width=0.32\textwidth]
    {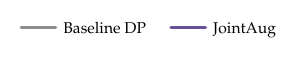}
    \vspace{-0.4cm}

    \caption{\textbf{Late-training coverage loss also occurs with
    \emph{low-dimensional state observations}.}
    Across five tasks, baseline diffusion policies lose pass@$10$ at later
    checkpoints despite using no visual encoder. Joint $(o,a)$ augmentation
    preserves substantially more pass@$10$.}
    \label{fig:app_lowdim}
\end{figure*}

\textbf{The same trend appears without a visual encoder.}
Finally, we train diffusion policies directly from low-dimensional state observations. Across all five tasks in Fig.~\ref{fig:app_lowdim}, baseline pass@$10$ again declines at later checkpoints, while pass@$1$ remains stable or improves. Thus, the late-training coverage loss happens in the absence of any visual encoder. 

Together, these experiments show that visual-encoder overfitting cannot account for most of the late training coverage loss. Freezing the encoder does not eliminate the decline, a newly initialized denoiser reproduces it under a frozen representation, and the same trend appears in low dimensional state-based policies with no visual encoder at all. The visual encoder can still affect the magnitude of the effect,
but the peak-then-decline behavior arises from denoiser training itself.

This result matches the generalization--memorization mechanism identified in prior work on diffusion models. With finite training data, the empirical denoising objective is minimized by the empirical score rather than the population score; and prolonged optimization can drive the denoiser toward memorizing training examples\citep{gu2023memorization,baptista2025memorization,ye2026provable}. Network training initially favors the smoother, population-like components of the score and fits the higher frequency, dataset-specific components later, particularly at low noise \citep{bonnaire2026diffusion,Favero2025BiggerIA}. Our results show the same transition in conditional diffusion policies and it persists after removing visual-representation learning and is concentrated in
the denoising network.

\appsubsec{Best-of-$k$ Off-Manifold Error Reveals the Same Coverage--Reliability Tradeoff}
\label{app:manifold_adherence}
\begin{figure}[t]
    \centering
    \includegraphics[width=0.4\columnwidth]
    {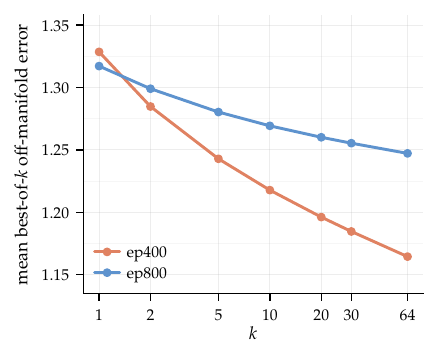}
    \caption{\textbf{Best-of-$k$ off-manifold error reveals the same coverage--reliability tradeoff.}
    At $k=1$, ep800 has slightly lower off-manifold error;
    from $k=2$ onward, ep400 is better and the gap grows with $k$.}
    \label{fig:app_muchado_offmanifold}
\end{figure}

We additionally ask whether the coverage--reliability tradeoff appears under a different offline metric of policy quality. Following the off-manifold metric of \citet{pan2026much}, we perturb expert actions at high-ambiguity training states, execute the perturbed actions to obtain OOD observations, and measure the residual of each predicted action from the span of expert actions at nearby training states. We extend from the one policy prediction sample to best-of-$k$ over stochastic policy samples.

We compare epochs 400 and 800 of \texttt{3-Piece Assembly} run. The later checkpoint has higher pass@$1$ ($0.328$ vs.\ $0.2988$) but lower pass@$10$ ($0.591$ vs.\ $0.639$). For each of 400 critical states, we generate 16 action-perturbed OOD observations. At each observation, we draw 64 policy action chunks and compute the expected minimum off-manifold error among $k$ samples; $k=1$ recovers the single-sample metric originally introduced by the paper.

Fig.~\ref{fig:app_muchado_offmanifold} shows the same tradeoff in this metric. At $k=1$, the later checkpoint has slightly lower off-manifold error. However, the ordering reverses at $k=2$, and the gap increases with $k$. Thus, the later checkpoint produces a slightly better typical sample under the original single-sample metric, whereas the earlier checkpoint enjoys a substantially better favorable tail. Mean off-manifold error therefore does not reveal the coverage–reliability tradeoff across training checkpoints. It may be more useful for comparing different policy methods under matched training budget, as in the original study that introduced the metric. Here we show that best-of-$k$ off-manifold error reveals the same coverage--reliability tradeoff observed with held-out action error and closed-loop pass@$k$.

\appsubsec{How Policy Coverage Affects World-Model Learning}
\label{app:world_model_coverage}

We additionally test whether policy coverage affects world-model learning from policy rollouts. For each task, we construct two training sets that share the same 200 expert demonstrations and differ only in the policy used to collect 1,000 additional unfiltered rollout episodes. One uses an earlier, higher-coverage checkpoint selected by the coverage-reliability criterion and the other uses the final, lower-coverage checkpoint. We keep failed rollouts because they provide additional state--action--next-state outcomes for learning dynamics and counterfactual reasoning on unseen actions. 

We train the same latent dynamics model on each dataset where images are encoded once with the same frozen DINOv3 ViT-B/16 encoder~\citep{simeoni2025dinov3}, so the comparison is not affected by representation learning. Each model receives two frames of history and predicts the next three latent states open loop using the predicted actions. We evaluate both models on the same held-out episodes collected across the full checkpoint family.

\begin{table*}[t]
    \centering
    \begin{minipage}[t]{0.47\textwidth}
        \centering
        \textbf{Threading (D2)}
        \vspace{0.3em}

        \resizebox{\linewidth}{!}{%
        \begin{tabular}{lccc}
            \toprule
            Training rollouts & $h=1$ & $h=2$ & $h=3$ \\
            \midrule
            Selected & \textbf{0.0800} & \textbf{0.1033} & \textbf{0.1186} \\
            Final    & 0.0815 & 0.1059 & 0.1219 \\
            \bottomrule
        \end{tabular}
        }
    \end{minipage}
    \hfill
    \begin{minipage}[t]{0.47\textwidth}
        \centering
        \textbf{Mug Cleanup (D1)}
        \vspace{0.3em}

        \resizebox{\linewidth}{!}{%
        \begin{tabular}{lccc}
            \toprule
            Training rollouts & $h=1$ & $h=2$ & $h=3$ \\
            \midrule
            Selected & \textbf{0.0992} & \textbf{0.1275} & \textbf{0.1458} \\
            Final    & 0.1003 & 0.1291 & 0.1484 \\
            \bottomrule
        \end{tabular}
        }
    \end{minipage}

    \caption{\textbf{Higher-coverage rollout data improves held-out world-model prediction.}
    We report open-loop latent token MSE; lower is better. Each model uses the same 200 demonstrations plus 1,000 unfiltered
    rollouts from either the earlier selected or final policy checkpoint.}
    \label{tab:world_model_coverage}
\end{table*}

\textbf{Higher-coverage rollout data produces better held-out prediction.}
Tab.~\ref{tab:world_model_coverage} shows lower prediction error for the early-rollout model at every horizon on both tasks. 
These results mirror the critic-learning experiment. Under a fixed rollout budget, broader policy coverage provides a wider range of state--action outcomes for learning. Because the rollouts are unfiltered, this includes both
successful and failed behavior; both are useful for learning how different actions lead to different future states.



\section{Implementation Details}
\label{sec:implementation_details}

\appsubsec{Tasks Datasets, and Evaluation}
\label{app:tasks_and_datasets}
Across both MimicGen and RoboCasa365, we use the demonstration datasets released with the official repositories. Unless otherwise noted, experiments in the main paper use image observations as policy inputs.

For MimicGen, we train policies on 200 demonstration trajectories for the main-text experiments. We additionally report results with 500 demonstrations in Appendix~\ref{app:coverage_reliability_full_results}. For RoboCasa pretraining, we use 65 atomic tasks with 100 human demonstrations per task. For SFT, we use 100 demonstrations per target task.

For MimicGen BC evaluation, pass@$k$ is computed over 500 held-out task configurations. For residual RL finetuning, we report the training success rate measured as the fraction of successful episodes among the 100 most recently completed episodes.

For MimicGen BC evaluation, pass@$k$ is computed over 500 held-out task configurations. For each configuration we run $K$ independent rollouts and record the number of successes $c$; for every $k \le K$ we compute $\mathrm{pass}@k = 1 - \binom{K-c}{k}/\binom{K}{k}$, i.e.\ the fraction of the $\binom{K}{k}$ $k$-subsets of the $K$ rollouts that contain at least one success, averaged over configurations. Best-of-$k$ action error is computed analogously: for each held-out state we sample $K$ action chunks, sort their errors $e_{(1)} \le \dots \le e_{(K)}$, and compute $\sum_{r=1}^{K-k+1} \frac{\binom{K-r}{k-1}}{\binom{K}{k}}\, e_{(r)}$, the expected minimum error over a uniformly random $k$-subset of the $K$ draws, averaged over states. Instead of using the first $k$ passes, we use use all $K$ samples to approximate both quantities, which gives the same expectation but reduces variance. For residual RL finetuning, we report the training success rate measured as the fraction of successful episodes among the 100 most recently completed episodes.

For RoboCasa evaluation, pass@$k$ is computed over 50 held-out task configurations for each evaluated task.

For both task suites, we use the default task horizon specified by the official repos. 

\appsubsec{Baseline Single Task Diffusion Policy}
\label{app:impl_dp}

Our baseline single-task BC policy is a CNN-based diffusion policy following the implementation of~\cite{chi2025diffusion}: a 1D temporal U-Net denoiser conditioned on a global observation embedding via FiLM, with a ResNet-18 visual encoder trained from scratch jointly with the denoiser network.
The policy consumes a single observation step (two $84{\times}84$ RGB views, stacked as a
$6$-channel input, plus the $9$-d proprioceptive state) and predicts an $8$-step action chunk with $6$-d rotation representation; all $8$ predicted actions are executed open-loop before the next inference. Training uses the standard denoising objective: a uniformly sampled diffusion step $t \in \{0,\dots,99\}$, forward noising with a squared-cosine $\beta$ schedule, and an MSE loss on the predicted noise ($\epsilon$-prediction). At inference we sample with $10$ DDIM steps ($\eta = 0$). All six MimicGen tasks share the identical architecture and optimization settings listed in Tab.~\ref{tab:app_dp_hparams}.

\begin{table}[h]
\centering
\captionsetup{skip=6pt}
\caption{Baseline diffusion-policy configuration. All settings are shared across the six MimicGen
tasks unless noted.}
\label{tab:app_dp_hparams}
\small
\begin{tabular}{ll}
\toprule
\textbf{Hyperparameter} & \textbf{Value} \\
\midrule
Observation steps & 1 \\
RGB inputs & 2 views, $84{\times}84$ (6-channel stack) \\
Proprioceptive state & 9-d \\
Predicted action chunk size & 8 \\
Executed action chunk size & 8 \\
\midrule
\multicolumn{2}{l}{\textit{Visual encoder}} \\
Backbone & ResNet-18 \\
Normalization & GroupNorm \\
Encoder sharing & one encoder per camera (not shared) \\
Image augmentation & random crop to $76{\times}76$ (center crop at inference) \\
\midrule
\multicolumn{2}{l}{\textit{Denoiser}} \\
Architecture & 1D conditional U-Net \\
Channel dimensions & $[512, 1024, 2048]$ \\
Kernel size / GroupNorm groups & 5 / 8 \\
Diffusion step embedding dim & 128 \\
\midrule
\multicolumn{2}{l}{\textit{Diffusion process}} \\
Training diffusion steps & 100 \\
$\beta$ schedule & squared-cosine ($\beta \in [10^{-4}, 0.02]$) \\
Prediction type / loss & $\epsilon$-prediction \\
Inference sampler & DDIM, 10 steps, $\eta = 0$ \\
\midrule
\multicolumn{2}{l}{\textit{Optimization}} \\
Optimizer & AdamW ($\beta = (0.95, 0.999)$, $\epsilon = 10^{-8}$) \\
Learning rate & $1\times 10^{-4}$ \\
Weight decay & $1\times 10^{-6}$ \\
LR schedule & cosine, 500 warmup steps \\
Batch size & 512 \\
Training epochs & 800 \\
Demonstrations & 200 (500 for Three-Piece Assembly D2 only) \\
EMA & inv\_gamma 1.0, power 0.75, $\max$ decay 0.9999 \\
Evaluation weights & EMA \\
\bottomrule
\end{tabular}
\end{table}

\appsubsec{\JointOA{} Implementation}
\label{app:impl_jointaug}

\JointOA{} augments policy training with samples from a joint generative model of the
full observation--action tuple $(o, a)$, trained on the same demonstrations as the policy. The
generator is a pixel-space, dual-output stream diffusion model that denoises the
$6{\times}84{\times}84$ two-camera frame, the $9$-d proprioceptive states together with the $8{\times}10$
action chunk under a single shared diffusion process.
Image tokens and the low-dimensional tokens are concatenated, jointly self-attend, and are split
again, so every modality is generated conditioned on the others rather than independently. All
three modalities are noised at the same timestep and the model predicts the $v$-target for each;
the loss is a weighted sum of per-modality mean squared errors with equal weights, which
prevents the image from drowning out the $80$-d action.

\paragraph{Inference.} Samples are drawn with the generator's EMA weights using a DDIM sampler
with $100$ steps. Training uses $1000$-step squared-cosine schedule. 

\textbf{Selecting the generator checkpoint window.}
We save generator checkpoints every 50 epochs. For each saved checkpoint $G_\gamma$, we draw 1024 joint samples $(\tilde o,\tilde a)\sim G_\gamma$ and compare each generated action chunk with two equally sized sets of expert action chunks: the demonstrations used to train the generator, $\mathcal{D}_{\mathrm{train}}$, and a held-out set $\mathcal{D}_{\mathrm{held}}$ of the same task that are used to train neither the generator nor the policy. Let $d(\tilde a,\mathcal{D})=\min_{a'\in\mathcal{D}}\|\tilde a-a'\|_2$ be the nearest-neighbor distance on normalized action chunks. We define
\begin{equation}
\rho(\gamma)=\frac{\mathbb{E}_{\tilde a\sim G_\gamma}\!\left[d(\tilde a,\mathcal{D}_{\mathrm{train}})\right]}{\mathbb{E}_{\tilde a\sim G_\gamma}\!\left[d(\tilde a,\mathcal{D}_{\mathrm{held}})\right]}.
\label{eq:gen_nn_ratio}
\end{equation}
$\rho(\gamma)=1$ means generated actions are equally close to training and held-out demonstrations. We set $\gamma_0$ to the first saved checkpoint with $\rho(\gamma)<1$, when generated samples first become closer to the training data than to held-out data, and sample from $\gamma_0$ and the four subsequent checkpoints spaced 100 epochs apart. Table~\ref{tab:app_jointaug_runs} lists the resulting checkpoints for each task.

\paragraph{Distillation into policy training.} During policy training, 50\% of
minibatch's rows is replaced by freshly generated $(o, a)$ tuples; replaced and dataset rows are
then trained with the policy's ordinary behavior-cloning loss, so there's no change to the policy architecture of
Appendix~\ref{app:impl_dp}. 

\begin{table}[h]
\centering
\captionsetup{skip=6pt}
\caption{Joint $(o,a)$ generator configuration. All settings are shared across the six MimicGen
tasks unless noted.}
\label{tab:app_jointaug_gen}
\small
\begin{tabular}{ll}
\toprule
\textbf{Hyperparameter} & \textbf{Value} \\
\midrule
Image & 2 views, $84{\times}84$ (6-channel stack) \\
Proprioceptive state & 9-d \\
Action chunk & 8 steps, absolute pose, 6-d rotation (10-d) \\
Observation steps & 1 \\
\midrule
\multicolumn{2}{l}{\textit{Image stream}} \\
Architecture & conv U-Net \\
Channel dimensions & $[96, 192, 384]$ \\
Bottleneck & self-attention at $21{\times}21$ \\
\midrule
\multicolumn{2}{l}{\textit{Low-dimensional stream}} \\
Architecture & DiT over 9 tokens (1 state + 8 action) \\
Model width / layers / heads & 384 / 6 / 6 \\
Dropout & 0.1 \\
\midrule
\multicolumn{2}{l}{\textit{Cross-modal coupling}} \\
Bridge & MM-DiT block at the bottleneck (joint self-attention) \\
Time embedding dim & 256 \\
\midrule
\multicolumn{2}{l}{\textit{Diffusion process}} \\
Training diffusion steps & 1000 \\
$\beta$ schedule & squared-cosine ($\beta \in [10^{-4}, 0.02]$) \\
Prediction type & $v$-prediction \\
Loss & per-modality mean MSE, weights $1{:}1{:}1$ (image:state:action) \\
Sampler & DDIM, 100 steps \\
Sampling weights & EMA \\
\midrule
\multicolumn{2}{l}{\textit{Optimization}} \\
Optimizer & AdamW ($\beta = (0.95, 0.999)$, $\epsilon = 10^{-8}$) \\
Learning rate & $1\times 10^{-4}$ \\
Weight decay & $1\times 10^{-4}$ \\
LR schedule & cosine \\
Batch size & 256 \\
EMA & inv\_gamma 1.0, power 0.75, $\max$ decay 0.9999 \\
Demonstrations & 200 (500 for Three-Piece Assembly D2) \\
\bottomrule
\end{tabular}
\end{table}

\begin{table}[h]
\centering
\captionsetup{skip=6pt}
\caption{Per-task \JointOA{} settings. Generator checkpoints are indexed by their training
epoch; rows of each minibatch are drawn equally from the listed checkpoints}
\label{tab:app_jointaug_runs}
\footnotesize
\setlength{\tabcolsep}{4pt}
\begin{tabular}{lcccccc}
\toprule
 & \textbf{Mug} & \textbf{Square} & \textbf{Threading} & \textbf{Hammer}
 & \textbf{3-Piece} & \textbf{3-Piece} \\
 & \textbf{D1} & \textbf{D2} & \textbf{D2} & \textbf{D1}
 & \textbf{D1} & \textbf{D2} \\
\midrule
Demonstrations & 200 & 200 & 200 & 200 & 200 & 500 \\
\midrule
Generator checkpoints & 100, 200,   & 100, 200,   & 100, 200, & 100, 200, & 100, 200, & 50, 150, \\
\quad (training epoch)  & 300, 400, & 300, 400   & 300, 400, & 300, 400, & 300, 400, & 250, 350, \\
                        & 500       &  500       & 500       & 500       & 500       & 450 \\
\bottomrule
\end{tabular}
\end{table}

\appsubsec{\PostBC{} Implementation}
\label{app:impl_postbc}
We follow the image instantiation of Algorithms~1 and~4 (Appendix~D of \cite{wagenmaker2025posterior}). An
ensemble of $K$ members is first fit to the policy's own training split; each member $f_l$ is a
freshly initialized copy of the policy's observation encoder followed by a 3-layer MLP head that
regresses the flattened $8{\times}10$ action chunk by mean squared error, and each is fit on its
own bootstrap resample of that split. The ensemble is trained once per task,
before and independently of policy training, with the settings in
Table~\ref{tab:app_postbc_ensemble}.

Posterior uncertainty is estimated from the disagreement of the ensemble predictions, cached per
training window offline. At every fetch we draw a fresh noise vector from the diagonal of that covariance, scale it by $\alpha$, and add it to the clean target action with the gripper dimension zeroed; policy training is otherwise unchanged from Appendix~\ref{app:impl_dp} or~\ref{app:impl_robocasa}. For MimicGen tasks, we use $\alpha = 2.0$. For RoboCasa, we use $\alpha = 1.0$.

\begin{table}[h]
\centering
\captionsetup{skip=6pt}
\caption{\PostBC{} posterior-ensemble config and noise injection. All settings are shared
across the five MimicGen tasks for which we report \PostBC{}.}
\label{tab:app_postbc_ensemble}
\small
\begin{tabular}{ll}
\toprule
\textbf{Hyperparameter} & \textbf{Value} \\
\midrule
\multicolumn{2}{l}{\textit{Ensemble members}} \\
Ensemble size $K$ & 20 \\
Member architecture & policy observation encoder + MLP head \\
Head width / layers & 512 / 3 \\
Member output & flattened action chunk ($8 \times 10 = 80$-d) \\
\midrule
\multicolumn{2}{l}{\textit{Ensemble optimization}} \\
Optimizer & Adam \\
Learning rate & $3\times 10^{-4}$ \\
Batch size (per member) & 128 \\
Training epochs & 200 \\
Checkpoing used & 40 \\
\midrule
\multicolumn{2}{l}{\textit{Noise injection}} \\
Noise strength $\alpha$ & 2.0 \\
Excluded dimensions & gripper (final action dimension, scale 0) \\
\bottomrule
\end{tabular}
\end{table}

\appsubsec{Context-Smoothed Pre-training in \TMRL{} Implementation}
\label{app:impl_tmrl}
Context-smoothed pre-training \cite{hong2026tmrl} widens a BC policy's support by corrupting its conditioning during pretraining, so that one network models $p(a \mid \tilde{c}, t_c)$ from conditional action distribution at $t_c = 0$ to the marginal action distribution at $t_c = 1$. We keep the paper's corruption kernel and noise schedule: a discrete linear $\beta$ schedule over $[10^{-4}, 0.02]$ with $T_c = 100$ context steps, the variance-preserving marginal $\tilde{c} = \sqrt{\bar{\alpha}_{t_c}}\, c + \sqrt{1 - \bar{\alpha}_{t_c}}\, \epsilon$, and $t_c \sim \mathrm{Unif}(0, 1)$ drawn independently of the action timestep $t_a$. We evaluate
with a clean context, $t_c = 0$.

\appsubsec{Residual RL Implementation}
\label{app:impl_dice_rl}

We use DICE-RL~\citep{sun2026prior} as an instantiation of residual RL finetuning. In DICE-RL, a frozen pretrained diffusion policy is finetuned by additive correction from a residual actor. Let $s$ be the observation embedding of the frozen policy and $z$ the initial DDIM noise; the base policy deterministically maps $(s,z)$ to a chunk $\pi_{\text{pre}}(s,z)$ of $H=8$ actions, and the executed chunk is $a = \pi_{\text{pre}}(s,z) + r_\theta(s,z)$, where only the residual MLP $r_\theta$ is trained. Since the chunk is executed open-loop, one transition spans the entire chunk and the critic $Q_\phi(s,a)$ scores the whole chunk. 
Because image features learned by BC on MimicGen tasks carry a weak value signal, we parameterize the residual actor and the critic with the low-dimensional state; the frozen pretrained policy still remains image-based.
Episodes end at the first success step or at the task horizon, with truncation treated as terminal.
Hyperparameters are identical across tasks (Table~\ref{tab:app_dice_rl_hparams}). For each residual RL experiment, we run with three random seeds. 

\begin{table}[h]
\centering
\caption{Residual RL finetuning configuration, identical across tasks; only the frozen base
checkpoint and the environment horizon change.}
\label{tab:app_dice_rl_hparams}
\small
\begin{tabular}{ll}
\toprule
\textbf{Hyperparameter} & \textbf{Value} \\
\midrule
Frozen base policy & EMA weights, DDIM 10 steps ($\eta=0$), chunk $H=8$ \\
Residual actor & MLP $[1024,1024,1024]$, GELU, LayerNorm \\
Critic ensemble & $10\times$ MLP $[1024,1024,1024]$, GELU, LayerNorm \\
BC weight $\lambda_{\text{BC}}$ & 200  \\
TD loss / $n$-step & MSE / $n=3$ chunks \\
Discount $\gamma$ / target update $\tau$ & 0.99 / 0.01 \\
Parallel environments & 64 \\
Gradient steps per iteration & 40 \\
Batch size & 256 \\
RLPD expert ratio & $0.5 \rightarrow 0.1$ over the first 10k training steps \\
Optimizer & Adam, lr $1\times10^{-4}$, weight decay $1\times10^{-5}$ \\
Actor update frequency & 2 \\
BoN Exploration & $\arg\max_k \min_e Q$ over $K=16$ samples \\
\bottomrule
\end{tabular}
\end{table}

\appsubsec{Multitask Training on Robocasa}
\label{app:impl_robocasa}

We use the RoboCasa release of \texttt{openpi} and follow the RoboCasa $\pi_{0.5}$ benchmark
recipe unchanged, except for the pretraining learning rate (Table~\ref{tab:app_pi05_train}).

\paragraph{Pretraining.}
Starting from \texttt{pi05\_base}, the language model is frozen, the vision encoder is trained at
$0.1\times$ the base learning rate, and the action expert at the base learning rate, for 100k steps
at batch size 256 with AdamW, a 1{,}000-step warm-up, and cosine decay to $0.1\times$ the peak. The
benchmark uses a peak learning rate of $2.5\times10^{-5}$; we instead uses $1\times10^{-4}$
($4\times$ higher, decaying to $1\times10^{-5}$), which we find to work better, with the vision-encoder scaling unchanged.

\paragraph{Supervised fine-tuning.}
Each fine-tuning run initializes all weights from a pretraining checkpoint and uses the benchmark
fine-tuning recipe as is: peak learning rate $2.5\times10^{-5}$ with the same warm-up and cosine
schedule (50k steps), batch size 256, the same frozen/scaled parameter groups, and normalization
statistics recomputed on the SFT data. Runs differ only in their initialization; checkpoints are
saved every 2.5k steps and evaluated on the SFT tasks up to 20k steps.

\begin{table}[h]
\centering
\captionsetup{skip=6pt}
\caption{$\pi_{0.5}$ training configuration (pretraining / fine-tuning). All values follow the
RoboCasa $\pi_{0.5}$ benchmark except the pretraining peak learning rate (benchmark:
$2.5\times10^{-5}$).}
\label{tab:app_pi05_train}
\small
\begin{tabular}{ll}
\toprule
\textbf{Hyperparameter} & \textbf{Value} \\
\midrule
Peak learning rate & $\mathbf{1\times10^{-4}}$ / $2.5\times10^{-5}$ \\
Schedule & 1k-step warm-up, cosine decay to $0.1\times$ peak \\
Training steps & 100k / 50k \\
Global batch size & 256 \\
Optimizer & AdamW ($\beta_1{=}0.9$, $\beta_2{=}0.95$, wd $10^{-10}$), grad clip 1.0 \\
EMA decay (evaluated weights) & 0.99 \\
Action chunk / cameras / image size & 50 steps / 3 / $224{\times}224$ \\
\bottomrule
\end{tabular}
\end{table}

\appsubsec{Real Robot Experiments}
\label{app:real_robot_impl}

\textbf{BC Pretraining.}
We train a Diffusion Policy using one $224\times224$ image from each of the gripper-mounted main and ultrawide cameras and two timesteps of proprioceptive state, expressed relative to the current end-effector pose. Each camera has a separate CLIP-pretrained ViT-B/16 encoder with attention pooling, finetuned at $0.1\times$ the base learning rate. A 1D conditional U-Net predicts a 16-step action chunk at 20\,Hz (0.8\,s). Each 10D action contains relative end-effector translation (3D) and rotation (6D), plus gripper width. The diffusion process uses 50 training timesteps, a squared-cosine $\beta$ schedule, and $\epsilon$-prediction; inference uses 16 DDIM steps. We train for 208 epochs with AdamW ($\mathrm{lr}=7.5\times10^{-5}$, $\beta=(0.95,0.999)$, weight decay $10^{-6}$), batch size 64, cosine learning-rate decay with 5\% warmup, gradient clipping at 0.5, and weight EMA with decay 0.9999. We apply random cropping (95\%) and color jitter. Of the 740 demonstrations, 680 are used for training and 60 for held-out validation.

\begin{figure*}[t]
    \centering
    \includegraphics[width=\textwidth]{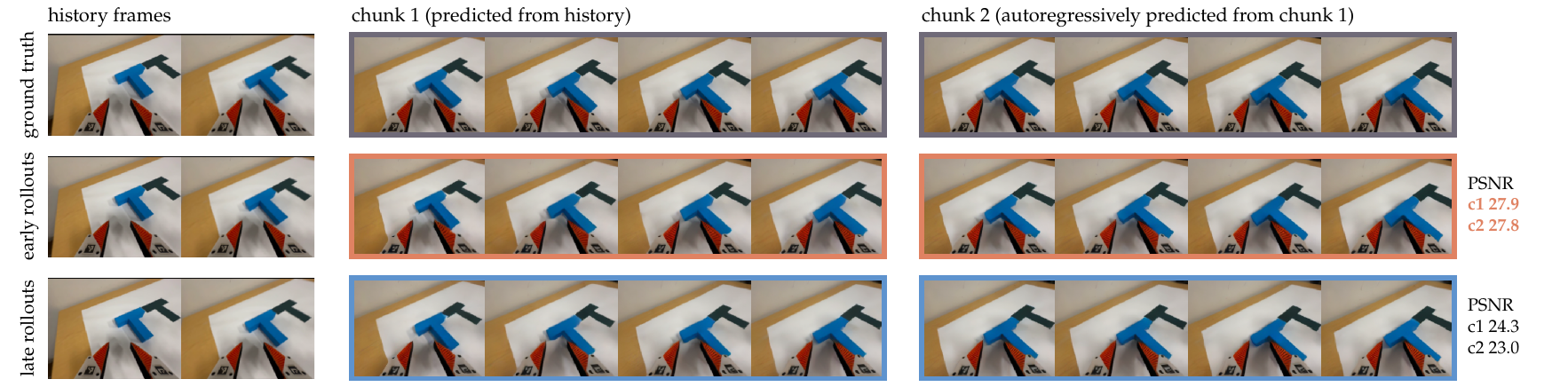}
    \vspace{-0.5cm}
    \caption{\textbf{Qualitative world-model predictions on real-robot Push-T.} The top row shows ground truth; the middle and bottom rows show predictions from world models trained on rollouts from the selected early and final policy checkpoints, respectively. Each model predicts four frames from two history frames (Chunk 1), then autoregressively predicts four more frames (Chunk 2). The early-rollout model achieves higher PSNR in both chunks.
    }
    \label{fig:wm_rollouts}
    \vspace{-0.4cm}
\end{figure*}

\textbf{World Model.}
The world model is a 12-layer diffusion transformer with width 768 and 12 attention heads. It operates on latents from a frozen Stable Diffusion VAE. Each training sequence contains six ultrawide-camera frames at 200\,ms intervals: two history frames and four target frames. We use causal attention and condition each predicted frame on the four actions executed since the previous frame by adding their embeddings to the frame tokens. We follow Diffusion Forcing and sample independent noise levels for each frame. Training uses $v$-prediction, a 1000-step squared-cosine schedule. Inference uses 50 DDIM steps. We use AdamW ($\mathrm{lr}=10^{-4}$, weight decay 0.01), 1000 warmup steps followed by a constant learning rate, batch size 32, and bf16 mixed precision. Fig.~\ref{fig:wm_rollouts} compares ground-truth frames (top row) with predictions from world models trained on early-checkpoint rollouts (middle row) and late-checkpoint rollouts (bottom row). The early-rollout model's predictions more closely match the ground truth.

\stopcontents[appendix]

\end{document}